\RequirePackage[svgnames,table]{xcolor}
\documentclass[11pt,letterpaper]{yalearxiv}

\usepackage[numbers,sort&compress]{natbib}
\usepackage{bm}
\usepackage{fontawesome5}
\usepackage{wrapfig}
\usepackage[ruled,vlined]{algorithm2e}
\usepackage{multirow}
\usepackage{tikz}

\ifdefined\XeTeXversion
  \microtypesetup{tracking=false}
\fi

\fancypagestyle{firststyle}{
  \fancyhf{}

  \fancyfoot[R]{\footerfont\thepage}
}

\usepackage{amsmath,amsfonts,bm}

\def\eqref#1{equation~\ref{#1}}

\def\1{\bm{1}}

\ifdefined\mathsfit\else
\DeclareMathAlphabet{\mathsfit}{\encodingdefault}{\sfdefault}{m}{sl}
\SetMathAlphabet{\mathsfit}{bold}{\encodingdefault}{\sfdefault}{bx}{n}
\fi

\newcommand{\AnchorLoopDynamicsFigure}[1]{%
  \begin{tikzpicture}
    \node[anchor=south west,inner sep=0] (basefigure) at (0,0)
      {\includegraphics[width=\linewidth]{#1}};
    \begin{scope}[x={(basefigure.south east)},y={(basefigure.north west)}]
      \fill[white] (0.2385,0.164) rectangle (0.3165,0.213);
      \fill[white] (0.5740,0.801) rectangle (0.6510,0.847);
      \fill[white] (0.9090,0.801) rectangle (0.9875,0.847);
      \pgfmathsetlengthmacro{\legendfontsize}{0.0096\linewidth}
      \tikzset{methodlegend/.style={anchor=west,inner sep=0,text=black,
        font=\sffamily\fontsize{\legendfontsize}{\legendfontsize}\selectfont}}
      \node[methodlegend] at (0.2400,0.186) {AnchorLoop (ours)};
      \node[methodlegend] at (0.5750,0.823) {AnchorLoop (ours)};
      \node[methodlegend] at (0.9100,0.823) {AnchorLoop (ours)};
    \end{scope}
  \end{tikzpicture}%
}

\tcbuselibrary{breakable}

\definecolor{tasksetterbg}{RGB}{220,233,244}
\definecolor{tasksetterfr}{RGB}{74,124,181}
\definecolor{solverbg}{RGB}{220,239,220}
\definecolor{solverfr}{RGB}{92,156,92}
\definecolor{criticbg}{RGB}{250,228,232}
\definecolor{criticfr}{RGB}{200,114,139}
\definecolor{tasksetterbg2}{RGB}{228,238,247}
\definecolor{solverbg2}{RGB}{232,244,230}
\definecolor{criticbg2}{RGB}{251,235,240}
\definecolor{perturbbg}{RGB}{240,240,240}
\definecolor{perturbfr}{RGB}{150,150,150}

\newtcolorbox{tasksetterbox}[1]{
  colback=tasksetterbg2,colframe=tasksetterfr,
  fonttitle=\bfseries,title=#1,breakable,
  arc=2pt,fontupper=\small
}
\newtcolorbox{solverbox}[1]{
  colback=solverbg2,colframe=solverfr,
  fonttitle=\bfseries,title=#1,breakable,
  arc=2pt,fontupper=\small
}
\newtcolorbox{criticbox}[1]{
  colback=criticbg2,colframe=criticfr,
  fonttitle=\bfseries,title=#1,breakable,
  arc=2pt,fontupper=\small
}
\newtcolorbox{perturbbox}[1]{
  colback=perturbbg,colframe=perturbfr,
  fonttitle=\bfseries,title=#1,breakable,
  arc=2pt,fontupper=\small
}

\newcommand{\cmark}{\textcolor{teal}{\ding{51}}}
\newcommand{\xmark}{\textcolor{red}{\ding{55}}}
\definecolor{verirow}{RGB}{255,228,228}

\providecommand{\Description}[2][]{}

\makeatletter
\def\munderbar#1{\underline{\sbox\tw@{$#1$}\dp\tw@\z@\box\tw@}}
\makeatother

\AddToHook{cmd/appendix/before}{%
  \setcounter{axiom}{0}%
}

\title{Self-Evolve With a Reference:\\Anchored Training of Tool-Integrated Agents}
\runningtitle{Self-Evolve With a Reference}
\date{7 October 2026}

\author[1]{Wenjie Liao}
\author[2]{Liangjie Zhao}
\author[2]{Zehong Cao}
\affil[1]{Waseda University\\
\texttt{jie3040@akane.waseda.jp}}
\affil[2]{Adelaide University\\
\texttt{zhaoliangjie55@gmail.com}, \texttt{Jimmy.Cao@adelaide.edu.au}}

\hypersetup{
  colorlinks=true,
  linkcolor=blue!50!black,
  citecolor=blue!50!black,
  urlcolor=blue!50!black,
  pdftitle={Self-Evolve With a Reference: Anchored Training of Tool-Integrated Agents},
  pdfauthor={Wenjie Liao, Liangjie Zhao, Zehong Cao}
}

\begin{document}

\begin{abstract}
Self-evolving tool-integrated agents learn from tasks and feedback generated within their own training loop. A Curriculum Agent generates tasks, while an Executor Agent learns from self-consistency signals through reinforcement learning. However, relying solely on the current Executor for feedback has two limitations: group-relative advantages vanish under full consensus, while uncertainty-based curriculum rewards favor disagreement without showing whether the generated tasks support further learning. These limitations motivate an additional reference beyond the current Executor. We propose \textit{AnchorLoop}, which introduces a frozen copy of the previous iteration's Executor as a historical reference and reuses it on both sides of the training loop. For the Executor, the anchor provides a cross-reference advantage that evaluates current outputs against both current and historical majority answers. For the Curriculum, it provides an agreement-based reference based on differences in sampled majority agreement. Since the Executor and anchor have identical parameters during Curriculum training, this comparison serves as a proxy for task selection rather than evidence of inter-version improvement or correctness. Across 13 reasoning benchmarks, AnchorLoop improves over Agent0 by 2.5\% on mathematical reasoning and 2.8\% on general reasoning tasks. It also maintains higher effective-advantage variance and continues improving in later iterations as the unanchored baseline shows diminishing gains. These results demonstrate the benefit of introducing a lightweight historical reference into self-evolving tool-integrated agents without external task or answer supervision.
\end{abstract}

\maketitle

\section{Introduction}
\label{sec:intro}

Tool-integrated language model agents can learn to solve complex tasks
through multi-turn interaction with external tools. Recent work further
extends this capability toward \emph{self-evolution}, where agents
generate not only solutions but also their own training tasks and
feedback. A representative paradigm couples a Curriculum Agent that
generates tasks with an Executor Agent that learns to solve them through
reinforcement learning \citep{xia2025agent0}. By generating tasks and
feedback within the training loop, such systems reduce their reliance
on externally supplied training tasks and answer labels. This creates a
closed learning loop in which the Executor serves not only as the
learner, but also as a source of feedback for training both agents.

This reliance on the current Executor, however, creates limitations on
both sides of self-evolution. On the Executor side, group-relative
learning relies on differences among sampled outputs. When all outputs
in a group reach consensus, the centered advantages vanish and provide
no relative preference among trajectories. On the Curriculum side,
uncertainty can identify tasks that elicit disagreement from the
Executor, but disagreement alone does not indicate whether those tasks
provide useful opportunities for further learning. In both cases, the
learning signal is determined entirely by the current Executor, without
a reference beyond its present behavior.

These limitations motivate a natural question: \emph{can self-evolution
gain an additional reference while remaining self-contained?} Our key
idea is to use the system's own recent history as that reference.
Specifically, we retain the Executor from the previous iteration as a
frozen \emph{anchor}. This introduces a temporal comparison without
requiring externally supplied tasks or answer labels. Historical
reference models have been used in iterative learning; for example,
SPIN compares against responses from a previous model while retaining
human reference responses \citep{chen2024spin}. Our focus is different:
we study how a recent Executor can serve as a shared reference for both
sides of a self-evolving Curriculum--Executor loop.

Based on this idea, we propose \textbf{AnchorLoop}, which reuses the
same historical anchor on both sides of self-evolution. For the
Executor, the anchor provides a \emph{cross-reference advantage} that
evaluates current outputs against both the current and anchor majority
answers, augmenting the preference induced by current self-consistency.
For the Curriculum, the anchor provides an agreement-based reference
through the positive difference in sampled majority-agreement rates.
The anchor is refreshed from the previous Executor at the beginning of
each outer iteration and remains frozen throughout that iteration.

Importantly, the two uses of the anchor have different interpretations.
Curriculum training precedes Executor training in each iteration, so
the current Executor and the newly refreshed anchor have identical
parameters during the Curriculum stage and produce independent rollout
groups. Their sampled agreement difference therefore serves as an
agreement-based proxy for shaping the Curriculum reward, rather than
direct evidence of improvement between model versions or correctness.
During subsequent Executor training, the current Executor evolves while
the anchor remains fixed, enabling a genuine historical comparison.
We further characterize this historical advantage and show that it acts
as a conditional reweighting of current outputs: it can introduce
additional relative information when current and historical majorities
differ, while still vanishing under full current consensus.

Our contributions are threefold:
\begin{itemize}
    \item \textbf{Historical Anchoring.}
    We introduce AnchorLoop, which uses the previous
    iteration's Executor as a shared historical reference for both
    Curriculum and Executor training.
    \item \textbf{Dual-Sided Learning.}
    We develop a cross-reference advantage for Executor training and
    an agreement-based reference for Curriculum training, and
    characterize how the historical reference shapes both learning
    signals.
    \item \textbf{Empirical Validation.}
    Across two backbones and thirteen reasoning benchmarks, AnchorLoop
    consistently improves over the unanchored baseline by 2.5-2.8\% points on average, while continuing to improve in later
    training iterations.
\end{itemize}
\section{Preliminaries}
\label{sec:prelim}

\noindent\textbf{Self-Evolving Curriculum--Executor Learning.}
We consider a self-evolving system consisting of a Curriculum Agent $\pi_\theta$ and an Executor Agent $\pi_\phi$. The Curriculum generates training tasks $x$, while the Executor learns to solve them through multi-turn interaction with external tools. Given a task $x$, the Executor generates a trajectory $\tau=(s_1,a_1,o_1,\ldots,s_H,a_H,o_H)$ of at most $H$ turns, where $a_h$ is either a tool call or a final answer and $o_h$ is the resulting observation. The trajectory terminates when the Executor produces a final answer $o(\tau)\in\mathcal{Y}$ or reaches the turn budget \citep{xue2025simpletir,xia2025agent0}. The Curriculum--Executor loop generates both training tasks and learning signals within the training process. The Curriculum evaluates generated tasks based on the current Executor's behavior, while the Executor learns from agreement among its own sampled outputs. Thus, both sides of self-evolution rely on feedback derived from the current Executor \citep{xia2025agent0}.

\noindent\textbf{Agreement-Based Group-Relative Learning.}
For each task $x$, the Executor samples $G$ trajectories
$\{\tau_i\}_{i=1}^{G}$. A fixed final-answer parser maps each
well-formed output to a canonical answer $o_i$. Following
self-consistency \citep{wang2023selfconsistency}, the unique modal
answer among valid outputs defines a pseudo-label $\tilde{y}$, and each
trajectory receives
\begin{equation}
    r_i=\mathbb{I}[o_i=\tilde{y}].
\end{equation}
Invalid or missing outputs receive zero reward and do not vote for
$\tilde{y}$, while groups without a unique modal answer are unresolved
and excluded from updates that require a pseudo-label.

\textbf{The reward $r_i$ measures \emph{agreement} rather than independently
verified \emph{correctness}}: the pseudo-label $\tilde{y}$ is determined
entirely by the sampled outputs. A consistent but incorrect group can
therefore receive the same agreement reward as a consistent correct
group.

Group-relative policy optimization (GRPO) \citep{shao2024grpo}
normalizes these rewards within the sampled group:
\begin{equation}
    A_i=\frac{r_i-\mu}{\sigma+\varepsilon},
\end{equation}
where $\mu$ and $\sigma$ denote the mean and standard deviation of the
group rewards, and $\varepsilon>0$ is a stabilizer. The policy is then
optimized using a clipped surrogate objective in the style of PPO
\citep{schulman2017ppo}. When all trajectories agree on the same valid
answer, $r_i=1$ for all $i$, yielding $A_i=0$. Thus, the centered
group-relative advantage provides no relative preference within a
fully consensual group.
\begin{figure*}[t]
    \centering
    \includegraphics[width=\textwidth]{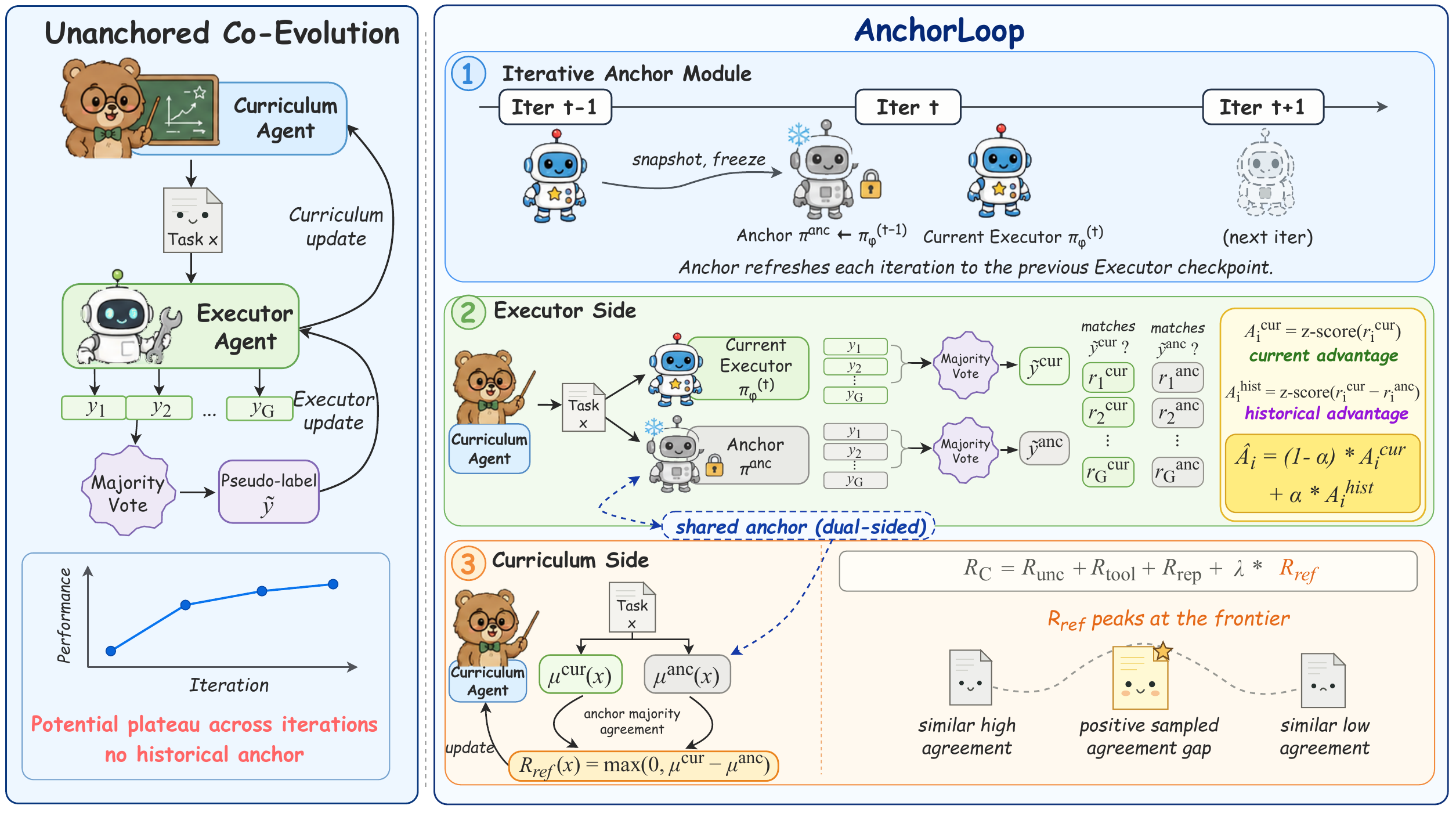}
    \caption{\textbf{Overview of AnchorLoop.}
    The left panel illustrates an unanchored two-agent self-evolution
    loop that can plateau. At outer iteration~$t$, AnchorLoop copies the
    previous Executor into a frozen anchor and reuses it on both sides
    of the loop. During Curriculum training, the current Executor and
    anchor have identical parameters but produce independent rollout
    groups for the agreement-based reference reward. All format-valid
    tasks form the same-iteration task pool $\mathcal{D}_t$. During
    Executor training, the anchor remains fixed while the current
    Executor is updated. Current groups are resampled at every inner
    step, whereas one anchor group per task and Curriculum batch is
    cached and reused to construct the cross-reference advantage.}
    \label{fig:framework}
    \Description{Overview of AnchorLoop. At each outer iteration, the
    previous Executor is frozen as an anchor and reused for both
    Curriculum and Executor training. The Curriculum compares agreement
    between independently sampled current and anchor rollout groups,
    while the Executor uses current and anchor majority answers to
    construct a cross-reference advantage.}
\end{figure*}

\section{Method}
\label{sec:method}

\subsection{Historical Anchor for Self-Evolution}
\label{sec:method:overview}

AnchorLoop introduces a historical reference into the
Curriculum--Executor loop defined in Section~\ref{sec:prelim}. At outer
iteration $t$, we retain the previous iteration's Executor as a frozen
anchor,
\begin{equation}
    \pi^{\mathrm{anc}} \leftarrow \pi_\phi^{(t-1)},
\end{equation}
and hold it fixed throughout the iteration. The anchor therefore
represents the most recently completed Executor while requiring no
externally supplied tasks or answer labels.

The same anchor is reused on both sides of self-evolution, but with
different roles. For Executor training, it provides a historical
majority answer used to construct a cross-reference advantage
(Section~\ref{sec:method:executor}). For Curriculum training, it
provides an additional rollout group whose sampled agreement is compared
with that of the current Executor
(Section~\ref{sec:method:curriculum}). The anchor is refreshed at the
beginning of every outer iteration, as illustrated in
Figure~\ref{fig:framework}.
Figure~\ref{fig:framework} is schematic, and
Equation~(\ref{eq:curriculum_reward}) defines the complete Curriculum reward.

The Curriculum and Executor are initialized from the same base model
$\pi_{\mathrm{base}}$ and updated separately. The Curriculum is
optimized with GRPO \citep{shao2024grpo}, while the Executor uses ADPO
\citep{xia2025agent0}. Curriculum training precedes Executor training
within each outer iteration. This update order gives the anchor
different interpretations across the two stages, as detailed in
Section~\ref{sec:method:training}.

\subsection{Executor: Cross-Reference Advantage}
\label{sec:method:executor}

The current agreement reward provides a relative preference only when
answers within a group differ. AnchorLoop augments this current-only
signal with an anchor-based comparison that changes how current outputs
are weighted. Because the anchor remains fixed throughout the
iteration, it provides a common historical reference as the Executor
updates. This design does not guarantee a nonzero advantage for every
task or prevent all forms of training stagnation.

\paragraph{Current and anchor references.}
For each task $x$ in the Executor stage, we sample $G$ trajectories
from the current Executor and another $G$ trajectories from the anchor.
Each resolved group follows the definition in
Section~\ref{sec:prelim} to produce a pseudo-label,
$\tilde{y}^{\mathrm{cur}}$ for the current group and
$\tilde{y}^{\mathrm{anc}}$ for the anchor group. If either group is
unresolved, the task is omitted from the pseudo-label-dependent
Executor update. Anchor rollouts are cached at the granularity of a
Curriculum batch and reused across the inner Executor steps that use
the same tasks, amortizing anchor sampling across those steps.

Let $o_i$ denote the parsed valid answer of current Executor trajectory
$i$. Invalid and missing trajectories receive zero for both indicators.
Each current trajectory receives two outcomes,
\begin{equation}
\begin{aligned}
    r_i^\mathrm{cur}
    &= \mathbb{I}[o_i=\tilde{y}^\mathrm{cur}], \\
    r_i^\mathrm{anc}
    &= \mathbb{I}[o_i=\tilde{y}^\mathrm{anc}].
\end{aligned}
\end{equation}
Both indicators score the same current answer $o_i$.
The first measures agreement with the current group's majority, whereas
the second measures agreement with the anchor group's majority rather
than scoring an anchor trajectory.

\paragraph{Cross-reference advantage.}
We form two advantages by centering and normalizing over the current
group:
\begin{equation}
\begin{aligned}
A_i^\mathrm{cur}
&=
\frac{r_i^\mathrm{cur}-\mu^\mathrm{cur}}
{\sigma^\mathrm{cur}+\varepsilon},
\\
A_i^\mathrm{hist}
&=
\frac{
(r_i^\mathrm{cur}-r_i^\mathrm{anc})-\bar{\Delta}
}{
s^\Delta+\varepsilon
},
\end{aligned}
\end{equation}
where $\mu^\mathrm{cur}$ and $\sigma^\mathrm{cur}$ are the mean and
standard deviation of $r^\mathrm{cur}$, while $\bar{\Delta}$ and
$s^\Delta$ are the mean and standard deviation of
$r^\mathrm{cur}-r^\mathrm{anc}$. The cross-reference advantage combines
the two:
\begin{equation}
    \hat{A}_i
    =
    (1-\alpha)A_i^\mathrm{cur}
    +\alpha A_i^\mathrm{hist},
    \qquad
    0\leq\alpha\leq1,
\end{equation}
where $\alpha$ controls the contribution of the historical advantage.

\paragraph{Properties of the historical term.}
The historical term acts as a conditional reweighting of current
outputs. For resolved current and anchor groups with
$\varepsilon>0$, several boundary cases clarify its behavior.
If the two majority labels agree, the two indicators coincide and
$A_i^\mathrm{hist}=0$. If all current answers are identical,
$r_i^\mathrm{cur}-r_i^\mathrm{anc}$ is constant across the current
group even when the majority labels differ. Centering therefore makes
both advantage terms zero. A different anchor label alone cannot
restore a relative preference within a fully consensual current group.

When the two majority labels differ and the current group contains
only those two answers, with both present, the historical advantage is
a positive rescaling of the current advantage. If the anchor's majority
answer is absent from the current group, the two advantages are
identical. With additional answer classes, the historical term can
distinguish outputs that match the anchor majority from other outputs
that disagree with the current majority. Its contribution therefore
depends on the sampled answer composition. These group-level relations
do not establish equivalence between complete training runs.

Refreshing the anchor ties its parameters to the most recently
completed Executor iteration. This does not establish that the anchor
remains informative on every task, nor that a reference fixed to the
base model must necessarily become uninformative.

\subsection{Curriculum: Anchor-Referenced Reward}
\label{sec:method:curriculum}

The Curriculum evaluates each generated task using sampled behavior
from the current Executor. AnchorLoop augments this current-only signal
with an independently sampled rollout group from the anchor. We first
define the inherited task-level Curriculum terms and then introduce the
anchor-referenced component.

\paragraph{Base Curriculum reward.}
Each candidate task $x$ is evaluated with $G$ current Executor
trajectories $\{\tau_j\}_{j=1}^{G}$. Let $n_y(x)$ denote the number of
valid parsed answers with canonical label $y$. We define
\begin{equation}
    p_{\max}(x)
    =
    \frac{1}{G}\max_y n_y(x),
    \qquad
    R_\mathrm{unc}(x)
    =
    1-2\left|p_{\max}(x)-\frac{1}{2}\right|.
\end{equation}
If no trajectory has a valid answer, we set $p_{\max}(x)=0$.
The maximum count remains defined when several labels tie, although
such a group has no pseudo-label. Invalid and missing trajectories do
not add to any label count and remain in the fixed denominator $G$.

Let $N_\mathrm{tool}(\tau_j)$ denote the number of Python calls executed
by the sandbox before trajectory $j$ terminates. Calls executed before
an invalid or missing final answer are still counted. With call cap
$C$, the task-level tool-use term is
\begin{equation}
    R_\mathrm{tool}(x)
    =
    \frac{1}{G}
    \sum_{j=1}^{G}
    \min\!\left(N_\mathrm{tool}(\tau_j),C\right).
\end{equation}

For a Curriculum batch $\mathcal{B}$ of $B$ candidate tasks, we compute
pairwise BLEU distances between task texts and connect two tasks when
their distance is below $d_\mathrm{rep}$. Let $\mathcal{C}(x)$ denote
the connected component containing $x$. The repetition term is
\begin{equation}
    R_\mathrm{rep}(x)
    =
    \frac{|\mathcal{C}(x)|}{B}.
\end{equation}
The binary gate $R_\mathrm{fmt}(x)$ equals one when the generated task
passes the fixed \texttt{Problem:} and \texttt{Answer:} format in
Appendix~\ref{app:prompt_curriculum}, and zero otherwise.

\paragraph{Anchor-referenced agreement.}
We augment these inherited terms with an anchor comparison. For each
candidate task $x$, we define
\begin{equation}
    R_\mathrm{ref}(x)
    =
    \max\!\bigl(
        0,\,
        \mu^\mathrm{cur}(x)-\mu^\mathrm{anc}(x)
    \bigr),
\end{equation}
where $\mu^\mathrm{cur}(x)$ and $\mu^\mathrm{anc}(x)$ are the majority
agreement rates of the current Executor and anchor, respectively.
For a resolved group, each rate is the number of rollouts matching its
own unique modal answer divided by the fixed $G$. Invalid and missing
outputs contribute zero to the numerator but remain in the denominator.
If either group is unresolved, we set $R_\mathrm{ref}(x)=0$; the
uncertainty, tool-use, repetition, and format terms remain defined.

Importantly, $\mu^\mathrm{anc}(x)$ is computed over anchor outputs and
is not the current-group mean of $r_i^\mathrm{anc}$ defined in
Section~\ref{sec:method:executor}. Moreover, the two agreement rates
are computed with respect to their respective group majorities rather
than a shared ground-truth answer.

\paragraph{Interpretation.}
The reference term measures a difference in sampled
\emph{agreement}, not a difference in \emph{correctness}. A positive
$R_\mathrm{ref}(x)$ means that the sampled current group has a larger
majority fraction than the sampled anchor group. It does not establish
that the current Executor is more accurate, nor that it can solve a
task that the anchor cannot. In particular, a more concentrated but
incorrect answer distribution can also receive a positive score.

The update order imposes a second limitation on this interpretation.
Because Curriculum training precedes Executor training, the current
Executor and the newly refreshed anchor have identical parameters
during the Curriculum stage and differ only through independent
sampling. Their sampled agreement rates may therefore differ, but such
a difference is not evidence of progress between historical model
versions. Taking the positive part does not remove this sampling
variation. We consequently use $R_\mathrm{ref}$ as an
agreement-based proxy for Curriculum shaping rather than as a measure
of correctness or inter-version improvement.

\paragraph{Complete Curriculum reward.}
We combine the reference term with the inherited uncertainty, tool-use,
repetition, and format-validity terms:
\begin{equation}
\label{eq:curriculum_reward}
\begin{aligned}
R_C(x)
=
R_\mathrm{fmt}(x)\max\!\big(&0,\,
w_\mathrm{unc}R_\mathrm{unc}(x)
+w_\mathrm{tool}R_\mathrm{tool}(x) \\
&-w_\mathrm{rep}R_\mathrm{rep}(x)
+\lambda R_\mathrm{ref}(x)
\big).
\end{aligned}
\end{equation}
The base reward weights used throughout the reported training runs are
listed in Table~\ref{tab:hyperparams}. When
$R_\mathrm{ref}(x)=0$, the remaining components of $R_C(x)$ remain
active; a zero reference term therefore does not by itself discard the
task.

\begin{algorithm}[t]
\SetAlgoLined
\DontPrintSemicolon
\caption{Training procedure of AnchorLoop.}
\label{alg:iagent0}
\textbf{Input:} Base model $\pi_{\mathrm{base}}$, outer iterations $T$\;
Initialize separate agents
$\pi_\theta,\pi_\phi\leftarrow\pi_{\mathrm{base}}$\;

\For{$t=1,\dots,T$}{
    Set
    $\pi^{\mathrm{anc}}\leftarrow\pi_\phi^{(t-1)}$
    and freeze the anchor for iteration $t$\;

    Keep $\pi_\phi$ fixed; update only $\pi_\theta$ via GRPO with
    reward $R_C$; collect all format-valid tasks in $\mathcal{D}_t$
    while retaining their Curriculum batch assignments\;

    Keep $\pi_\theta$ fixed; update only $\pi_\phi$ on
    $\mathcal{D}_t$ via ADPO using the scaled cross-reference
    advantage $\widetilde{A}$\;

    Resample each current group at every Executor inner step and reuse
    one cached anchor group per task and Curriculum batch\;

    Discard $\mathcal{D}_t$ and its rollout cache, and denote the
    resulting Executor by $\pi_\phi^{(t)}$\;
}
\textbf{Output:} Trained Executor $\pi_\phi^{(T)}$\;
\end{algorithm}

\subsection{Alternating Training with an Iterative Anchor}
\label{sec:method:training}

AnchorLoop alternates Curriculum and Executor training over $T$ outer
iterations. Let $\pi_\phi^{(t)}$ denote the Executor after iteration
$t$, with $\pi_\phi^{(0)}=\pi_{\mathrm{base}}$. At the beginning of
iteration $t$, we refresh the anchor as
\begin{equation}
    \pi^{\mathrm{anc}}
    \leftarrow
    \pi_\phi^{(t-1)}
\end{equation}
and keep it frozen throughout the iteration.

\paragraph{Curriculum stage.}
We first fix the Executor and update only the Curriculum Agent using
GRPO with the reward $R_C$ from
Section~\ref{sec:method:curriculum}. The current Executor and newly
refreshed anchor therefore have identical parameters
$\pi_\phi^{(t-1)}$ during this stage and differ only through independent
sampling.

All format-valid tasks generated during Curriculum training form the
same-iteration task pool $\mathcal{D}_t$, with their Curriculum batch
assignments preserved. We do not filter these tasks according to
$R_\mathrm{ref}$ or the total Curriculum reward $R_C$. The
Curriculum-generated \texttt{Answer:} field is used only for format
checking and is neither provided to the Executor nor used as a
supervision label.

\paragraph{Executor stage.}
We then freeze the Curriculum Agent and update only the Executor on
$\mathcal{D}_t$. The anchor remains fixed at
$\pi_\phi^{(t-1)}$, while the current Executor evolves during
optimization, providing the historical comparison used by the
cross-reference advantage in Section~\ref{sec:method:executor}.

Each task remains fixed across the Executor inner steps associated with
its Curriculum batch. For each task, the current Executor resamples
$G$ trajectories at every inner step, whereas the anchor group is
sampled once per task and Curriculum batch and cached across the
corresponding Executor updates. Current groups are not retained between
inner steps. After the Executor stage, the resulting policy is denoted
$\pi_\phi^{(t)}$. The task pool $\mathcal{D}_t$ and all cached rollouts
are then discarded. The next outer iteration generates a new task pool
and does not replay tasks from earlier iterations.

\paragraph{Policy optimization.}
For Curriculum training, the Curriculum Agent's group consists of a
batch of generated tasks. Each task is scored by $R_C$ from
Section~\ref{sec:method:curriculum}, and the resulting rewards are
normalized across the task batch to form the GRPO advantages.

For Executor training, each group consists of the $G$ trajectories
sampled for a single task. The standard group-relative advantage is
replaced by the cross-reference advantage $\hat{A}_i$ from
Section~\ref{sec:method:executor}. The Executor follows ADPO
\citep{xia2025agent0}, which scales the advantage according to current
answer agreement and dynamically adjusts the upper clipping bound.

For each task $x$, we compute the current-group agreement score
\begin{equation}
    \hat{p}(x)
    =
    \mu^\mathrm{cur}(x)
    =
    \frac{1}{G}
    \sum_{i=1}^{G}r_i^\mathrm{cur},
\end{equation}
where $\hat{p}(x)$ is the fraction of current outputs matching the
current group's modal answer. Let
$s(x)=f(\hat{p}(x))$ denote the ADPO task weight, where $f$ is
nondecreasing. We apply this weight to the complete cross-reference
advantage:
\begin{equation}
\begin{aligned}
    \widetilde{A}_i(x)
    &=s(x)\hat{A}_i \\
    &=s(x)\left[
        (1-\alpha)A_i^\mathrm{cur}
        +\alpha A_i^\mathrm{hist}
    \right].
\end{aligned}
\end{equation}
Thus, the same task weight scales both the current and historical
components. The clipping interval is
\[
    [1-\epsilon_\mathrm{low},\,
      1+\epsilon_\mathrm{high}(x)],
\]
where
$\epsilon_\mathrm{high}(x)=g(\hat{p}(x))$ and $g$ is nonincreasing.
Thus, both the task weight and the upper clipping bound are recomputed
from the current group's agreement score.

\paragraph{Token-level Executor objective.}
Let $y_{i,\ell}$ denote the token generated at position $\ell$ of
trajectory $i$, and let $h_{i,\ell}$ contain the task, preceding
generated tokens, and any tool observations returned before that
position. Tool observations condition subsequent generation but are
not themselves optimized. We use a mask $m_{i,\ell}$ over generated
reasoning, tool-call, and final-answer tokens, with
$M_i=\sum_\ell m_{i,\ell}$.

The behavior policy $\pi_{\phi_{\mathrm{old}}}$ is frozen before
sampling the current group at each Executor inner step. The token-level
importance ratio is
\begin{equation}
    \rho_{i,\ell}(\phi)
    =
    \frac{
        \pi_\phi(y_{i,\ell}\mid h_{i,\ell})
    }{
        \pi_{\phi_{\mathrm{old}}}(y_{i,\ell}\mid h_{i,\ell})
    }.
\end{equation}
The same trajectory-level advantage $\widetilde{A}_i(x)$ is applied to
all optimized tokens in trajectory $i$. The Executor objective is
\begin{equation}
\begin{aligned}
\mathcal{L}_\mathrm{exec}(\phi)
={}&
-\frac{1}{G}
\sum_{i=1}^{G}
\frac{1}{M_i}
\sum_\ell m_{i,\ell}
\min\!\Bigl(
\rho_{i,\ell}(\phi)\widetilde{A}_i(x),\\
&\qquad
\mathrm{clip}\!\left(
\rho_{i,\ell}(\phi),
1-\epsilon_\mathrm{low},
1+\epsilon_\mathrm{high}(x)
\right)
\widetilde{A}_i(x)
\Bigr)\\
&+
\frac{\beta_\mathrm{KL}}{G}
\sum_{i=1}^{G}
\frac{1}{M_i}
\sum_\ell m_{i,\ell}
D_\mathrm{KL}\!\left(
\pi_\phi(\cdot\mid h_{i,\ell})
\,\middle\|\,
\pi_\mathrm{ref}(\cdot\mid h_{i,\ell})
\right),
\end{aligned}
\end{equation}
where the KL reference is the frozen base model,
$\pi_\mathrm{ref}=\pi_\mathrm{base}$.

The anchor adds one frozen Executor checkpoint to storage at any given
time. Its rollouts are cached at the task--Curriculum-batch level,
amortizing anchor sampling across the corresponding Executor inner
steps. Overall, the anchor serves as an independently sampled reference
during Curriculum training and becomes a historical reference as the
Executor subsequently updates. Algorithm~\ref{alg:iagent0} summarizes
the complete training procedure, while the corresponding functions and
hyperparameters are reported in Appendix~\ref{app:hyperparams}.
Appendix~\ref{app:prompts} provides the Curriculum and Executor prompts,
together with the answer-parsing and group-resolution rules.

\section{Experiments and Results}
\label{sec:experiments}

\subsection{Experimental Setup}
\label{sec:exp:setup}

\paragraph{Training Details.}
We evaluate AnchorLoop on two open-source base models,
Qwen3-4B-Base \citep{yang2025qwen3} and MiMo-7B-Base
\citep{xiaomi2025mimo}. For each backbone, the Curriculum Agent,
Executor Agent, and initial anchor are derived from the same base model.
We implement AnchorLoop on VeRL \citep{sheng2024verl} with a sandboxed
Python interpreter provided by VeRL-Tool. A complete
\texttt{<python>}--\texttt{</python>} block triggers one subprocess
call, whose standard output is returned to the model as the next
observation. Training follows the alternating procedure in
Section~\ref{sec:method:training}. Each outer iteration consists of a
Curriculum stage followed by an Executor stage. All format-valid tasks
generated during Curriculum training enter the same-iteration Executor
task pool $\mathcal{D}_t$ while retaining their Curriculum batch
assignments; we apply neither reward-based task filtering nor task
replay across outer iterations. For each task, the current Executor
samples $G=8$ trajectories at every Executor inner step, whereas the
anchor samples $G=8$ trajectories once per task and Curriculum batch
and reuses them across the corresponding inner steps. The anchor is
refreshed from the previous iteration's Executor at the beginning of
each outer iteration and remains frozen throughout that iteration.

Unless otherwise specified, we report models after three outer
iterations. Table~\ref{tab:main_results} uses this setting for both backbones, while
the anchor-refresh analysis in Figure~\ref{fig:analysis}(a)
additionally extends training to a fourth iteration. AnchorLoop uses
no externally supplied training tasks or answer labels during
self-evolution; the Curriculum Agent generates the complete Executor
training task pool. Appendix~\ref{app:exp_details} provides the full
training configuration, hyperparameters, and sandbox settings.

\begin{table*}[t]
\centering
\caption{Results on mathematical and general reasoning benchmarks.
The two AVG columns summarize their respective benchmark groups. Best
per backbone is in \textbf{bold}.}
\label{tab:main_results}
\setlength{\tabcolsep}{1.2pt}
\renewcommand{\arraystretch}{1.08}
\footnotesize
\resizebox{\linewidth}{!}{%
\begin{tabular}{l c *{15}{c}}
\toprule
 & & \multicolumn{8}{c}{\textbf{Mathematical reasoning}} &
\multicolumn{7}{c}{\textbf{General reasoning}} \\
\cmidrule(lr){3-10}\cmidrule(l){11-17}
\textbf{Model Name} & \textbf{\faTools} & \textbf{AMC23} &
\textbf{MATH} & \textbf{GSM8K} & \textbf{Minerva} &
\textbf{Olympiad} & \textbf{AIME24} & \textbf{AIME25} &
\textbf{AVG} & \textbf{SuperGPQA} & \textbf{MMLU-Pro} &
\textbf{BBEH} & \textbf{GPQA-D} & \textbf{HumanEval} &
\textbf{AGIEval} & \textbf{AVG} \\
\midrule
\multicolumn{17}{l}{\textit{Qwen3-4B-Base}} \\
Base Model                  & \xmark & 45.4 & 68.6 & 88.5 & 37.3 & 41.2 & 11.0 & 6.33 & 42.6 & 21.1 & 37.6 &  7.6 & 36.5 & 65.7 & 54.8 & 37.2 \\
Base Model \textit{w/ tool} & \cmark & 45.9 & 72.7 & 88.8 & 38.1 & 42.4 & 12.5 & 7.86 & 44.0 & 25.8 & 43.6 &  8.5 & 37.2 & 67.1 & 55.4 & 39.6 \\
+ Absolute Zero             & \cmark & 50.3 & 76.8 & 88.8 & 40.6 & 42.9 & 12.7 & 13.0 & 46.4 & 26.4 & 52.9 &  8.7 & 38.4 & 68.4 & 56.7 & 41.9 \\
+ SPIRAL                    & \xmark & 56.7 & 77.8 & 90.1 & 42.8 & 39.6 & 13.5 & 10.2 & 47.2 & 26.5 & 53.0 &  9.4 & 38.9 & 67.9 & 57.3 & 42.1 \\
+ R-Zero                    & \xmark & 57.7 & 80.2 & 90.5 & 52.4 & 43.6 & 12.3 & 4.45 & 48.7 & 28.3 & 51.4 & 10.2 & 39.6 & 69.6 & 57.9 & 42.8 \\
+ Socratic-Zero             & \xmark & 57.7 & 80.3 & 88.5 & 47.3 & 49.2 & 19.4 & 13.8 & 50.9 & 28.5 & 54.2 & 10.8 & 40.1 & 70.2 & 58.4 & 43.7 \\
+ Agent0                    & \cmark & 60.4 & 79.9 & 91.1 & 56.7 & 48.4 & 17.3 & 14.9 & 52.7 & 30.4 & 55.1 & 11.5 & 41.7 & 71.8 & 60.1 & 45.1 \\
\rowcolor{verirow}
+ \texttt{AnchorLoop} (ours) & \cmark & \textbf{63.5} & \textbf{82.9} & \textbf{94.3} & \textbf{58.1} & \textbf{49.7} & \textbf{20.5} & \textbf{17.4} & \textbf{55.2} & \textbf{32.7} & \textbf{58.7} & \textbf{14.3} & \textbf{44.3} & \textbf{74.5} & \textbf{62.8} & \textbf{47.9} \\
\midrule
\multicolumn{17}{l}{\textit{MiMo-7B-Base}} \\
Base Model                  & \xmark & 58.3 & 75.6 & 89.9 & 47.3 & 46.6 & 30.7 & 22.3 & 53.0 & 24.9 & 42.3 &  8.5 & 26.1 & 52.3 & 48.0 & 33.7 \\
Base Model \textit{w/ tool} & \cmark & 63.6 & 77.9 & 91.1 & 51.4 & 50.2 & 34.4 & 26.9 & 56.5 & 26.4 & 44.2 &  9.4 & 27.4 & 54.1 & 48.9 & 35.1 \\
+ Absolute Zero             & \cmark & 64.2 & 77.0 & 91.6 & 50.0 & 49.1 & 33.8 & 26.6 & 56.0 & 27.6 & 48.6 & 10.0 & 29.2 & 55.8 & 50.2 & 36.9 \\
+ SPIRAL                    & \xmark & 64.7 & 78.0 & 92.4 & 51.8 & 47.6 & 35.2 & 28.1 & 56.8 & 28.0 & 47.9 & 10.4 & 29.8 & 56.2 & 50.8 & 37.2 \\
+ R-Zero                    & \xmark & 65.0 & 80.5 & 93.3 & 57.2 & 51.3 & 33.0 & 28.6 & 58.4 & 28.8 & 48.4 & 10.7 & 31.0 & 57.0 & 51.5 & 37.9 \\
+ Socratic-Zero             & \xmark & 66.5 & 79.3 & 90.4 & 52.5 & 55.8 & 39.1 & 34.0 & 59.7 & 29.2 & 49.1 & 11.1 & 31.7 & 57.6 & 52.3 & 38.5 \\
+ Agent0                    & \cmark & 67.3 & 81.4 & 94.2 & 57.9 & 54.1 & 39.5 & 32.4 & 61.0 & 30.5 & 51.3 & 12.2 & 33.5 & 59.2 & 54.0 & 40.1 \\
\rowcolor{verirow}
+ \texttt{AnchorLoop} (ours) & \cmark & \textbf{70.0} & \textbf{84.1} & \textbf{95.7} & \textbf{60.6} & \textbf{56.7} & \textbf{42.3} & \textbf{35.4} & \textbf{63.5} & \textbf{33.0} & \textbf{54.0} & \textbf{14.6} & \textbf{36.4} & \textbf{62.1} & \textbf{56.7} & \textbf{42.8} \\
\bottomrule
\end{tabular}
}
\end{table*}

\paragraph{Baselines and Benchmarks.}
We compare against seven baselines: two no-training references
(Base Model and Base Model w/ Tool), four self-evolving methods
(Absolute Zero \citep{zhao2025absolutezero}, SPIRAL
\citep{liu2025spiral}, R-Zero \citep{huang2025rzero}, and
Socratic-Zero \citep{wang2025socratic}), and Agent0
\citep{xia2025agent0}, the direct unanchored baseline underlying our
two-agent training setup.

We evaluate mathematical reasoning on seven benchmarks: AMC23 \citep{maa_amc}, MATH-500
\citep{hendrycks2021math,lightman2023verify}, GSM8K
\citep{cobbe2021gsm8k}, Minerva \citep{lewkowycz2022minerva},
OlympiadBench \citep{he2024olympiad}, AIME24 \citep{maa2024aime1,maa2024aime2}, and AIME25 \citep{maa2025aime1,maa2025aime2}.
General reasoning is evaluated on six benchmarks: SuperGPQA
\citep{du2025supergpqa}, MMLU-Pro \citep{wang2024mmlupro}, BBEH
\citep{kazemi2025bbeh}, GPQA-Diamond \citep{rein2024gpqa}, HumanEval
\citep{chen2021humaneval}, and AGIEval \citep{zhong2024agieval}.
We report greedy pass@1 except mean@32 for AMC23 \citep{maa_amc} and AIME.
Appendix~\ref{app:baselines_benchmarks} provides baseline adaptations,
dataset details, and the complete evaluation and aggregation protocol.

\subsection{Main Results}
\label{sec:exp:main}

Table~\ref{tab:main_results} compares
AnchorLoop with the seven baselines across thirteen reasoning
benchmarks. Overall, historical anchoring consistently improves over
the unanchored Agent0 baseline on both backbones and across both
evaluation domains.

\paragraph{Mathematical reasoning.}
As shown in Table~\ref{tab:main_results}, AnchorLoop achieves a Math
AVG of $55.2$ on Qwen3-4B-Base and $63.5$ on MiMo-7B-Base, improving
over Agent0 by $2.5$ percentage points on both backbones. Relative to
the tool-augmented base models, the corresponding gains are $11.2$ and
$7.0$ points. The improvement is also consistent at the benchmark
level: AnchorLoop obtains the highest reported score in every
mathematical benchmark column on both backbones, including AIME24 and
AIME25.

\paragraph{General reasoning.}
Table~\ref{tab:main_results} shows a similar pattern outside the
mathematical domain. AnchorLoop reaches a General AVG of $47.9$ on
Qwen3-4B-Base and $42.8$ on MiMo-7B-Base, exceeding Agent0 by $2.8$
and $2.7$ percentage points, respectively. It also obtains the highest
reported score in every general-reasoning benchmark column on both
backbones. These improvements are notable because the Curriculum
generates mathematical training tasks, whereas the evaluation spans
knowledge-intensive reasoning, coding, and general problem solving.
The results therefore show that the gains associated with historical
anchoring are not confined to the mathematical benchmarks used to
characterize the training distribution.

Appendix Table~\ref{tab:run_variability} reports the mean and sample
standard deviation over five independent training runs. These
statistics characterize run-to-run variability rather than statistical
significance. Appendix~\ref{app:additional} further reports
per-benchmark trajectories, iteration dynamics, and sensitivity results
on MiMo-7B-Base.

\subsection{Training Dynamics}
\label{sec:exp:dynamics}

\begin{figure*}[t]
\centering
\AnchorLoopDynamicsFigure{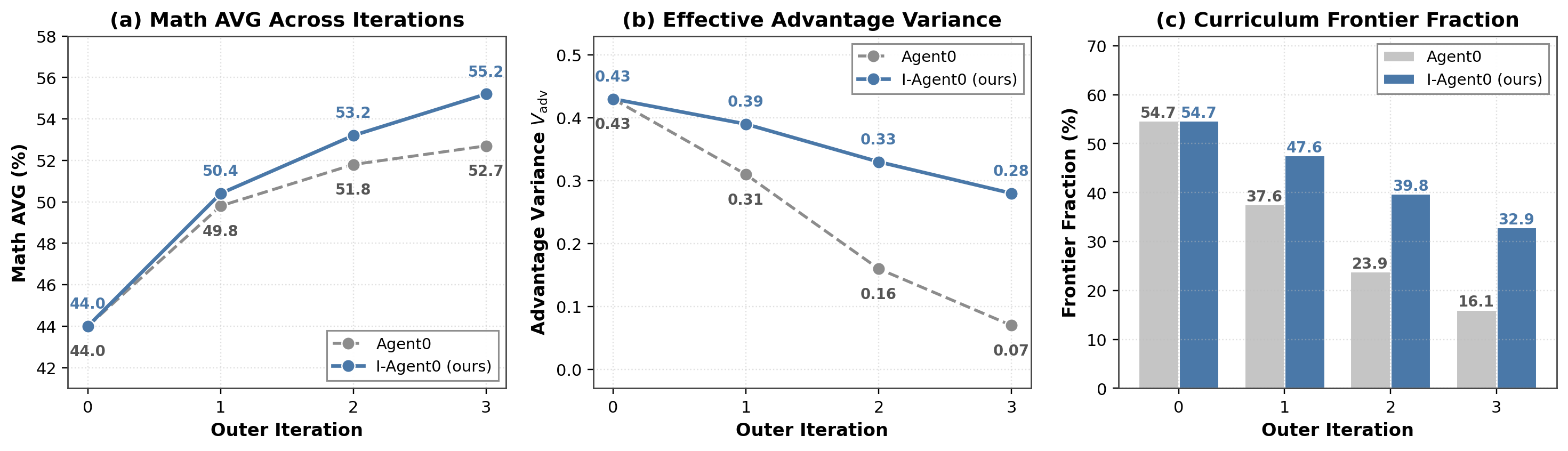}
\caption{\textbf{Training dynamics on Qwen3-4B-Base.}
(a) Math AVG across three outer iterations.
(b) Within-group variance of the effective Executor advantage,
$V_{\mathrm{adv}}$, averaged over sampled groups. By iteration~3,
it decreases to $0.07$ for Agent0 but remains at $0.28$ for
AnchorLoop.
(c) Fraction of Curriculum tasks with a positive sampled reference
gap. An unresolved current--reference pair contributes zero. The
fraction decreases from $54.7\%$ initially to $16.1\%$ for Agent0
and $32.9\%$ for AnchorLoop by iteration~3.}
\label{fig:dynamics}
\Description{Three panels compare Agent0 and AnchorLoop on
Qwen3-4B-Base across outer iterations using mathematical reasoning
performance, effective advantage variance, and the fraction of
Curriculum tasks with a positive sampled reference gap.}
\end{figure*}

\paragraph{Later-iteration performance.}
Figure~\ref{fig:dynamics}(a) tracks Math AVG across the three outer
iterations. Agent0 improves by $5.8$ points in the first iteration,
but its incremental gains decrease to $2.0$ and $0.9$ points in
iterations~2 and~3. AnchorLoop begins from the same initialization and
shows a comparable early improvement, while retaining gains of $2.8$
and $2.0$ points in the subsequent iterations. Consequently, the gap
between the two methods grows across later training and reaches $2.5$
points after iteration~3. This pattern motivates examining whether the
learning signals available to the two methods evolve differently over
training.

\paragraph{Effective advantage variance.}
Figure~\ref{fig:dynamics}(b) measures the within-group variance of the
final effective Executor advantage. Let $\mathcal{B}_t$ contain all
sampled Executor groups across the update steps of outer iteration $t$.
For group size $G$ and final ADPO-scaled advantage
$\widetilde{A}_{b,i}$, we compute
\[
V_{\mathrm{adv}}^{(t)}
=
\frac{1}{|\mathcal{B}_t|}
\sum_{b\in\mathcal{B}_t}
\left[
\frac{1}{G}
\sum_{i=1}^{G}
\left(
\widetilde{A}_{b,i}
-
\frac{1}{G}
\sum_{j=1}^{G}\widetilde{A}_{b,j}
\right)^2
\right].
\]
Both methods begin at $V_{\mathrm{adv}}=0.43$. By iteration~3,
Agent0 decreases to $0.07$, whereas AnchorLoop retains substantially
more within-group variation at $0.28$. This observation is consistent
with the motivation for cross-reference training: the anchored
objective preserves more differentiated effective advantages in the
sampled groups as training proceeds. It does not imply that the
historical component is nonzero for every group; as established in
Section~\ref{sec:method:executor}, both components vanish when all
current answers are identical. Nor does this statistic alone establish
that the increased variance causes the performance improvement.

\paragraph{Positive reference-gap frequency.}
Figure~\ref{fig:dynamics}(c) examines the Curriculum side by reporting
the fraction of generated tasks for which the independently sampled
current and reference groups satisfy
\[
    \mu^\mathrm{cur}(x)>\mu^\mathrm{ref}(x).
\]
An unresolved group pair contributes zero and is not counted as
positive. For AnchorLoop, the reference group is sampled from the
newly refreshed anchor; for Agent0, we construct the diagnostic using
a second independent rollout group from the same Executor checkpoint.
During Curriculum training, the current and reference models have
identical parameters, and each agreement rate is computed against its
own group's majority answer. The measured fraction decreases from
$54.7\%$ initially to $16.1\%$ for Agent0 and $32.9\%$ for AnchorLoop
by iteration~3.

Consistent with the interpretation in
Section~\ref{sec:method:curriculum}, this statistic measures the
frequency of positive sampled agreement differences rather than
correctness or inter-version progress. The higher fraction observed
for AnchorLoop therefore characterizes the sampled Curriculum signal;
it does not imply that the current Executor has newly solved those
tasks or establish that the reference reward causes the subsequent
performance difference.

\subsection{Anchor Analysis}
\label{subsec:ablation}

\begin{table}[t]
\centering
\caption{\textbf{Tool comparisons and anchor ablations on Qwen3-4B-Base.}
Agent0 and AnchorLoop are shared endpoints. Only the anchor variants
use matched training and evaluation settings.}
\label{tab:tool_ablation}
\small
\setlength{\tabcolsep}{2pt}
\renewcommand{\arraystretch}{1.05}
\begin{tabular}{lcc}
\toprule
\textbf{Method} & \textbf{Math AVG} & \textbf{General AVG} \\
\midrule
\multicolumn{3}{l}{\textit{Without tool access}} \\
Qwen3-4B (Base)                   & 42.6 & 37.2 \\
$\quad +\,$ SPIRAL                & 47.2 & 42.1 \\
$\quad +\,$ R-Zero                & 48.7 & 42.8 \\
\midrule
\multicolumn{3}{l}{\textit{With tool access}} \\
$\quad +\,$ TIR                   & 44.0 & 39.6 \\
$\quad +\,$ Absolute Zero         & 46.4 & 41.9 \\
\midrule
\multicolumn{3}{l}{\textit{Matched anchor variants with tool access}} \\
Agent0                              & 52.7 & 45.1 \\
$\quad +\,R_{\mathrm{ref}}$ only    & 53.6 & 45.9 \\
$\quad +\,A^{\mathrm{hist}}$ only   & 54.4 & 46.7 \\
\rowcolor{verirow}
Full \texttt{AnchorLoop}            & \textbf{55.2} & \textbf{47.9} \\
\bottomrule
\end{tabular}
\end{table}

\paragraph{Contribution of the two anchor signals.}
The matched anchor block of Table~\ref{tab:tool_ablation} separates the two uses of the anchor:
$R_{\mathrm{ref}}$ on the Curriculum side and
$A^{\mathrm{hist}}$ on the Executor side. All four configurations use
the same Qwen3-4B-Base initialization, three outer iterations, update
steps, ADPO settings, tool environment, and evaluation protocol. When
the historical advantage is disabled, the Executor uses
$A^\mathrm{cur}$ with unit coefficient. When enabled, it uses
$(1-\alpha)A^\mathrm{cur}+\alpha A^\mathrm{hist}$ with
$\alpha=0.6$. Disabling the Curriculum reference sets $\lambda=0$
while leaving all other Curriculum rewards and task handling unchanged.

Adding $R_{\mathrm{ref}}$ alone improves the Math and General averages
from $52.7/45.1$ to $53.6/45.9$, while adding
$A^{\mathrm{hist}}$ alone reaches $54.4/46.7$. Combining both gives
the full AnchorLoop result of $55.2/47.9$. Thus, both anchor uses are
associated with improvements over the unanchored configuration in the
matched setting, with the Executor-side historical advantage producing
the larger individual change. The endpoint comparison does not,
however, isolate the mechanism behind either improvement or establish
a synergistic interaction between the two components.

\begin{figure*}[t]
\centering
\includegraphics[width=\linewidth]{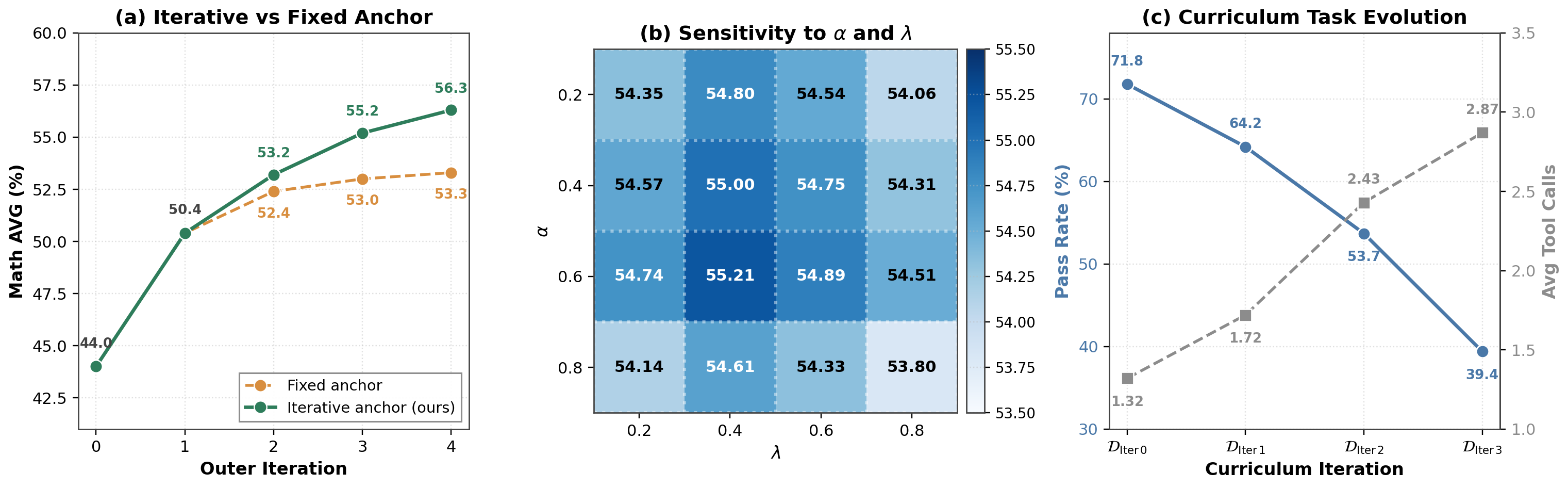}
\caption{\textbf{Anchor and Curriculum analyses on Qwen3-4B-Base.}
(a) Math AVG across outer iterations with an iterative anchor versus
an anchor fixed to the base model.
(b) Sensitivity over the historical-advantage weight $\alpha$ and
Curriculum reference weight $\lambda$ on a $4\times4$ grid; the
default setting is $(\alpha,\lambda)=(0.6,0.4)$.
(c) Pass rate and average tool calls on $200$ tasks sampled from each
Curriculum stage. The iteration-1 Executor generates $G=8$
trajectories per task, and both statistics are computed from the same
$1{,}600$ trajectories per stage.}
\label{fig:analysis}
\Description{Three panels analyze iterative versus fixed anchors,
sensitivity to the historical-advantage and Curriculum-reference
weights, and changes in sampled Curriculum tasks and tool use across
training.}
\end{figure*}

\paragraph{Iterative versus fixed anchor.}
Figure~\ref{fig:analysis}(a) tests whether refreshing the historical
reference matters by comparing AnchorLoop with an anchor permanently
fixed to the base model. The two settings remain close in the first
iteration, but diverge as training proceeds. The iterative and fixed
variants reach $55.2$ and $53.0$ at iteration~3, respectively, and
$56.3$ and $53.3$ at iteration~4. These results show that refreshing
the anchor with the most recently completed Executor is associated
with stronger later-iteration performance in this setting. They do
not establish that a fixed anchor necessarily becomes uninformative,
nor do they isolate checkpoint age as the sole source of the
difference.

\paragraph{Comparison across tool settings.}
Table~\ref{tab:tool_ablation} places AnchorLoop alongside methods evaluated with
and without tool access under their respective training configurations.
Adding the Python tool to the untrained Qwen3-4B base model changes
Math AVG from $42.6$ to $44.0$ and General AVG from $37.2$ to $39.6$.
Among the tool-integrated training methods, AnchorLoop reaches
$55.2/47.9$, compared with $52.7/45.1$ for Agent0.

These comparisons provide context for tool-integrated performance but
should not be interpreted as a controlled ablation of tool access.
The methods differ simultaneously in task generation, training
objective, and reward design. The comparisons across methods in
Table~\ref{tab:tool_ablation} therefore do not
isolate the causal contribution of either tool use or historical
anchoring. Appendix~\ref{app:anchor_analysis} further analyzes the
historical-advantage structure, computational cost, anchor refresh and
caching, and representative failure cases.

\subsection{Curriculum Analysis}
\label{sec:exp:sensitivity}
\label{sec:exp:curriculum}

\paragraph{Sensitivity to $\alpha$ and $\lambda$.}
Figure~\ref{fig:analysis}(b) evaluates the two anchor-specific weights:
the historical-advantage weight $\alpha$, which balances
$A^\mathrm{cur}$ and $A^\mathrm{hist}$ in Executor training, and the
Curriculum reference weight $\lambda$, which scales
$R_\mathrm{ref}$. Across the $4\times4$ grid, the default setting
$(\alpha,\lambda)=(0.6,0.4)$ reaches a Math AVG of $55.21$, while all
evaluated configurations remain within $1.5$ points of this value.
The result indicates limited sensitivity within the evaluated range,
although it does not characterize settings outside the grid or remove
the need to select the two weights.

\paragraph{Evolution of Curriculum tasks and tool use.}
We further examine how the generated task distribution changes across
Curriculum stages $t=0,1,2,3$. From each stage, we sample $200$ tasks
and evaluate them using a fixed iteration-1 Executor, which
independently generates $G=8$ trajectories per task. We compute the
mean binary agreement reward and the average number of tool calls over
the same $1{,}600$ trajectories for each stage.

Figure~\ref{fig:analysis}(c) shows that the measured pass rate under
this fixed Executor decreases from $71.8\%$ on
$\mathcal{D}_{\mathrm{Iter}\,0}$ to $39.4\%$ on
$\mathcal{D}_{\mathrm{Iter}\,3}$, while the average number of tool
calls per trajectory increases from $1.32$ to $2.87$. Because the same
Executor is used to evaluate tasks from all stages, these changes
indicate that the distribution of Curriculum-generated tasks evolves
over training: later task sets lower agreement reward and more
tool interactions from this fixed evaluator.

These statistics should not be interpreted as direct measures of task
difficulty or correctness. In particular, $R_\mathrm{ref}$ compares
independently sampled agreement rates during Curriculum training rather
than identifying tasks that the current Executor can solve but the
anchor cannot. The observed trends therefore characterize how the
generated task and response distributions change alongside anchored
self-evolution; they do not by themselves establish that later tasks
are intrinsically harder, that additional tool calls are necessary, or
that $R_\mathrm{ref}$ causally produces either trend.
Appendix~\ref{app:case} provides selected Curriculum tasks,
post-update Executor comparisons, and full reasoning traces.

\section{Conclusion}
\label{sec:conclusion}

We introduce AnchorLoop to address weakening feedback in self-evolving Curriculum and Executor training. At each outer iteration, a frozen copy of the previous Executor provides an independently sampled agreement reference for Curriculum training and helps construct a cross-reference advantage for Executor training, without external tasks or answer labels. Across two backbones and thirteen benchmarks, AnchorLoop improves Math AVG over Agent0 by 2.5 points and General AVG by 2.7 to 2.8 points. It also retains higher effective-advantage variance and shows stronger gains in later iterations.

\FloatBarrier
\bibliographystyle{plainnat}
\bibliography{iclr2026_conference}

\clearpage

\appendix
\section{Experimental Details}
\label{app:exp_details}

\subsection{Training Setup}
\label{app:train_setup}

We implement AnchorLoop on top of VeRL \citep{sheng2024verl} with the
VeRL-Tool extension providing a sandboxed Python interpreter that
executes each code block emitted by the Executor in a separate
subprocess. All experiments use 8 NVIDIA A100 80GB GPUs. For
AnchorLoop, one outer iteration takes about 24 hours on
Qwen3-4B-Base \citep{yang2025qwen3} and about 36 hours on
MiMo-7B-Base \citep{xiaomi2025mimo}. Multiplying these per-iteration
values gives approximate three-iteration references of 72 and 108
hours. These figures are estimates rather than separately measured
total times. Within each outer iteration, all tasks that pass the fixed
Curriculum output format are retained with their original batch
assignments and passed to the Executor stage. We do not select tasks by $R_\mathrm{ref}$, the total reward $R_C$, or a top-$k$ rule. The Executor prompt contains the problem statement and any required context.
The Curriculum-generated Answer: field is used only to check
the generation format and is not shown to the Executor or used as a
training label.

A task remains fixed across the Executor inner steps assigned to its
Curriculum batch. The current Executor resamples $G=8$ trajectories at
every inner step. The anchor samples $G=8$ trajectories once per task
and batch, then reuses them across those steps. This amortizes anchor
sampling while keeping current rollouts aligned with the changing
Executor. The task pool, current groups, and anchor cache are discarded
after the Executor stage. The next outer iteration uses newly generated
tasks without replaying earlier tasks.

\subsection{Hyperparameters}
\label{app:hyperparams}

Table~\ref{tab:hyperparams} summarizes the confirmed settings for both
backbones. AnchorLoop use ADPO functions and
clipping configuration on Qwen3-4B-Base and MiMo-7B-Base. The learning
rate is lower for the larger MiMo backbone. The
weight $\alpha = 0.6$ on the historical advantage and the weight
$\lambda = 0.4$ on the reference reward are the defaults selected by
the sweeps.

\paragraph{Curriculum reward configuration.}
The Executor group size is $G=8$, and the Curriculum batch size is
$B=64$. The tool-use term caps each trajectory at $C=4$ executed calls.
The repetition graph uses BLEU distance threshold
$d_\mathrm{rep}=0.6$. The format gate and nonnegative truncation apply
to the complete weighted sum. AnchorLoop use
$w_\mathrm{unc}=1.0$, $w_\mathrm{tool}=0.25$, and
$w_\mathrm{rep}=0.20$ on both backbones. These weights remain fixed
across all outer iterations.

\paragraph{ADPO configuration.}
AnchorLoop configuration follows the functional form of Agent0's released
ADPO implementation \citep{xia2025agent0}. For an agreement
score $p$, let $p_{\min}$ and $p_{\max}$ delimit the interpolation
interval. We first map $p$ to a value in $[0,1]$,
\begin{equation}
q(p)=\min\!\left\{1,\max\!\left\{0,
\frac{p-p_{\min}}{p_{\max}-p_{\min}}\right\}\right\}.
\end{equation}
Let $s_{\min}$ be the minimum advantage weight, $\epsilon_0$ the
base clipping width, and $\delta_\epsilon$ the maximum increment
to the upper width. The functions in Section~\ref{sec:method}
are instantiated as
\begin{equation}
\begin{aligned}
f(p)&=s_{\min}+(1-s_{\min})q(p),\\
g(p)&=\epsilon_0+\delta_\epsilon[1-q(p)],\\
\epsilon_\mathrm{low}&=\epsilon_0.
\end{aligned}
\end{equation}
With the listed values, $f(p)$ ranges from $0.6$ to $1.0$, and
$g(p)$ ranges from $0.3$ at low agreement to $0.2$ at high agreement.
The Executor's clipping interval is therefore $[0.8,1+g(p)]$.
These functions saturate outside the interpolation interval and do
not by themselves filter tasks. The score $p$ is recomputed from
each current Executor group, and $f(p)$ scales the complete mixed
advantage. No additional normalization is applied to this task weight.

The base
width $\epsilon_0=0.2$ is also the Executor's lower clipping width,
and the Curriculum uses the fixed width $\epsilon_C=0.2$. Its clipping
interval is $[1-\epsilon_C,1+\epsilon_C]$. At each Executor inner step,
the rollout policy is frozen as $\pi_{\phi_\mathrm{old}}$ before the
current group is sampled. Token-level PPO ratios use this behavior
policy. The KL term instead uses the frozen base model
$\pi_\mathrm{ref}=\pi_\mathrm{base}$ on both backbones.

\begin{table}[h]
\centering
\caption{Training settings for both backbones. The base Curriculum reward weights are shared by baselines and AnchorLoop.}
\label{tab:hyperparams}
\small
\setlength{\tabcolsep}{6pt}
\renewcommand{\arraystretch}{1.1}
\begin{tabular}{lll}
\toprule
\textbf{Group} & \textbf{Hyperparameter} & \textbf{Value} \\
\midrule
\multirow{4}{*}{Outer loop}
 & Outer iterations $T$            & 3 \\
 & Group size $G$                  & 8 \\
 & Curriculum inner steps $K_C$    & 32 \\
 & Executor inner steps $K_E$      & 128 \\
\midrule
\multirow{2}{*}{Anchor side}
 & Historical advantage weight $\alpha$ & 0.6 \\
 & Frontier reward weight $\lambda$ & 0.4 \\
\midrule
\multirow{4}{*}{Optimization}
 & Learning rate (Qwen3-4B)        & $1\times10^{-6}$ \\
 & Learning rate (MiMo-7B)         & $5\times10^{-7}$ \\
 & KL coefficient $\beta_{\mathrm{KL}}$ & 0.01 \\
 & KL reference $\pi_\mathrm{ref}$ & Frozen base model \\
\midrule
\multirow{6}{*}{Curriculum}
 & Clip width $\epsilon_C$ & 0.2 \\
 & Uncertainty weight $w_\mathrm{unc}$ & 1.0 \\
 & Tool-use weight $w_\mathrm{tool}$ & 0.25 \\
 & Repetition weight $w_\mathrm{rep}$ & 0.20 \\
 & Tool-call cap $C$ & 4 \\
 & BLEU distance threshold $d_\mathrm{rep}$ & 0.6 \\
\midrule
\multirow{5}{*}{ADPO}
 & Lower score limit $p_{\min}$ & 0.3 \\
 & Upper score limit $p_{\max}$ & 0.8 \\
 & Minimum weight $s_{\min}$ & 0.6 \\
 & Base clip width $\epsilon_0$ & 0.2 \\
 & Upper width increment $\delta_\epsilon$ & 0.1 \\
\midrule
\multirow{3}{*}{Rollout}
 & Curriculum batch size           & 64 tasks \\
 & Max turns $H$                   & 8 \\
 & Max output tokens per turn      & 4096 \\
\midrule
\multirow{3}{*}{Sandbox}
 & Code execution timeout          & 30 s \\
 & Memory cap per call             & 1 GB \\
 & Max code block length           & 1024 chars \\
\bottomrule
\end{tabular}
\end{table}

\subsection{Sandbox Configuration}
\label{app:sandbox}

The Python interpreter runs in a stateless subprocess with an
execution timeout of 30 seconds and a memory limit of 1 GB per call.
The standard scientific stack is preinstalled, including NumPy,
SciPy, SymPy, and the standard math, fractions, and itertools
modules. A tool call must be enclosed by one complete
\texttt{<python>} and \texttt{</python>} tag pair. Each complete pair
triggers one sandbox call. An incomplete or malformed pair does not
trigger execution. The interpreter's standard output is returned to
the Executor as the next observation. The tool-use analysis reports the configurations
and evaluation settings shown in the main paper. It does not provide a
matched same-checkpoint comparison with the Python tool disabled.

\section{Prompt Templates}
\label{app:prompts}

We present the prompt templates used by the Curriculum Agent and the
Executor Agent. Both prompts are kept short and remain unchanged
throughout the three outer iterations. The iterative anchor shapes
the policy distribution through the reward and the advantage, not
through prompt edits.

\subsection{Curriculum Agent Prompt}
\label{app:prompt_curriculum}

The Curriculum Agent generates one math problem per call. The prompt
asks for a self-contained problem with a single numerical answer that
the Executor can attempt through the tool interface defined above.

\begin{tcolorbox}[colback=gray!5, colframe=gray!40,
  title=\textbf{Curriculum Agent Prompt}, fonttitle=\bfseries,
  fontupper=\small]
You are a math problem creator. Generate one challenging mathematical
reasoning problem that satisfies the requirements below.\\[4pt]
1. The problem requires careful reasoning across multiple steps, not
pattern matching or memorization.\\
2. The problem has a single unambiguous numerical answer that can be
verified by computation.\\
3. The problem is self-contained. All needed information appears in
the problem text.\\
4. The problem can be approached with high school or early
undergraduate mathematics, including algebra, number theory,
combinatorics, geometry, or calculus.\\
5. A Python interpreter is available to the solver for verifying
intermediate computations.\\[4pt]
Output the problem and the correct answer in the following format.\\[4pt]
\texttt{Problem: <problem statement>}\\
\texttt{Answer: <numerical answer>}
\end{tcolorbox}

\subsection{Executor Agent Prompt}
\label{app:prompt_executor}

The Executor Agent receives the problem statement and produces a
trajectory that interleaves reasoning with Python tool calls. The
prompt instructs the Executor to wrap each code block in one complete
\texttt{<python>} and \texttt{</python>} pair. It also requires a final
answer in a fixed format that the consensus rule and binary reward can
parse.

\begin{tcolorbox}[colback=gray!5, colframe=gray!40,
  title=\textbf{Executor Agent Prompt}, fonttitle=\bfseries,
  fontupper=\small]
You are a math problem solver with access to a Python interpreter.\\[4pt]
Solve the following problem step by step. At any point during your
reasoning you may write a short Python script and execute it by
enclosing the code in \texttt{<python>} and \texttt{</python>} tags.
The standard output of the script will be returned to you as the
next observation, and you may continue reasoning from there.\\[4pt]
When you have determined the final answer, output it on its own line
in the format\\[4pt]
\texttt{Final Answer: <numerical answer>}\\[4pt]
\texttt{Problem:}\\
\texttt{\{problem\}}
\end{tcolorbox}

\subsection{Answer Parsing and Group Resolution}
\label{app:answer_protocol}

We parse the fixed final-answer field before comparing sampled outputs.
A well-formed output is mapped to one canonical answer label. An output
is invalid when the trajectory finishes but the field is absent,
malformed, or contains conflicting answers. A trajectory is missing
when it reaches a length limit, times out, terminates with an execution
failure, or returns empty output before a final answer. Invalid and missing outputs do not vote for a modal label and receive reward zero, but they remain in the fixed group
size $G=8$.

A group is resolved only when its valid answers contain one label with
a unique highest count. A highest-count tie or the absence of a valid
answer makes the group unresolved. Such a group is omitted from an
Executor update that requires a pseudo-label. In a diagnostic average
over a fixed sampled pool, its pseudo-label-dependent contribution is
zero, so the stated denominator does not change. If either group in the
Curriculum comparison is unresolved, its frontier term is set to zero
while the uncertainty, tool-use, and repetition terms remain active.
The same parsing and resolution rules are used for current, anchor, and
reference groups.

\section{Baselines and Benchmark Details}
\label{app:baselines_benchmarks}

\subsection{Baseline Methods}
\label{app:baselines}

We compare AnchorLoop against seven baselines that cover both untrained
references and published self-evolving training methods.

\textbf{Base Model.} The pre-trained Qwen3-4B-Base or MiMo-7B-Base
backbone evaluated without any further training and without access to
the Python interpreter.

\textbf{Base Model w/ Tool (TIR).} The same backbone with access to
the Python interpreter at inference time but no additional training.

\textbf{Absolute Zero \citep{zhao2025absolutezero}.} A zero-data
framework in which the model proposes code reasoning tasks and solves
them through Python execution. It is trained without external tasks or
labels. In our adaptation, the Python interpreter is available during
both training and evaluation.

\textbf{SPIRAL \citep{liu2025spiral}.} A self-play reinforcement
learning method that trains a shared policy through multi-turn
self-play in TextArena zero-sum games.

\textbf{R-Zero \citep{huang2025rzero}.} A zero-data reasoning
framework that improves the base model through iterative training
data the model generates itself, without a code executor.

\textbf{Socratic-Zero \citep{wang2025socratic}.} A self-evolving
approach that starts from 100 seed problems. It trains a Solver and a
Generator under guidance from a fixed, high-capacity Teacher, without
a Python code executor.

\textbf{Agent0 \citep{xia2025agent0}.} Agent0 trains a Curriculum
Agent to propose tasks and an Executor Agent to solve them through
Python tool use. The Agent0 baseline retains Ambiguity-Dynamic Policy
Optimization (ADPO), which adjusts advantage weights and clipping
bounds according to answer agreement. AnchorLoop retains both ADPO
components and introduces an iterative anchor into the Executor
advantage and the Curriculum reward.

All baselines that involve training use the same Qwen3-4B-Base and
MiMo-7B-Base backbones and the same evaluation protocol as AnchorLoop.
All results in tables
were obtained from evaluations conducted in this work.

\subsection{Evaluation Benchmarks}
\label{app:benchmarks}

Table~\ref{tab:benchmarks} lists seven mathematical and six general
reasoning benchmarks. Agent0 and AnchorLoop scores average five
independent training runs for each method on each backbone. Domain averages are unweighted means over benchmarks. They are computed from
unrounded benchmark scores and then rounded to one decimal place. For AMC23, AIME24, and AIME25,
mean@32 averages correctness over 32 responses per question and then
across questions. Other benchmarks use greedy pass@1.

For the Curriculum-stage analysis in Figure~\ref{fig:analysis}(c), we
sample $200$ tasks from each stage. The iteration-$1$ Executor
independently generates $G=8$ trajectories per task, yielding $1{,}600$
trajectories per stage. Pass rate is the mean binary reward across these
trajectories. Average tool calls is the arithmetic mean of the number of
Python calls over the same trajectories.

\begin{table}[t]
\centering
\caption{Run-level variability for the most recent baseline (Agent0) and AnchorLoop. Entries report
mean $\pm$ sample standard deviation over five independent training runs.}
\label{tab:run_variability}
\small
\setlength{\tabcolsep}{4pt}
\renewcommand{\arraystretch}{1.05}
\begin{tabular}{llcc}
\toprule
\textbf{Backbone} & \textbf{Domain} & \textbf{Agent0} & \textbf{AnchorLoop} \\
\midrule
Qwen3-4B & Math    & $52.7 \pm 0.68$ & $55.2 \pm 0.51$ \\
Qwen3-4B & General & $45.1 \pm 0.57$ & $47.9 \pm 0.44$ \\
MiMo-7B  & Math    & $61.0 \pm 0.63$ & $63.5 \pm 0.49$ \\
MiMo-7B  & General & $40.1 \pm 0.55$ & $42.8 \pm 0.46$ \\
\bottomrule
\end{tabular}
\end{table}

\subsubsection{Mathematical Reasoning}

\textbf{AMC23.} Problems from the 2023 American Mathematics
Competitions, covering high school algebra, geometry,
number theory, and combinatorics. \citep{maa_amc}

\textbf{MATH-500 \citep{hendrycks2021math,lightman2023verify}.} A subset of $500$
challenging problems from the MATH dataset, spanning algebra,
counting and probability, geometry, intermediate algebra, number
theory, prealgebra, and precalculus.

\textbf{GSM8K \citep{cobbe2021gsm8k}.} A dataset of grade school math
word problems that require arithmetic across multiple steps together
with basic algebra.

\textbf{Minerva (OCWCourses) \citep{lewkowycz2022minerva}.} This
benchmark contains $272$ problems from undergraduate STEM courses in
MIT OpenCourseWare.

\textbf{OlympiadBench \citep{he2024olympiad}.} Math problems from
international mathematical olympiads and similar competitions.

\textbf{AIME24 and AIME25.} The 2024 \citep{maa2024aime1,maa2024aime2} and 2025 \citep{maa2025aime1,maa2025aime2} editions of the
American Invitational Mathematics Examination, with thirty problems
each that cover advanced high school mathematics.

\subsubsection{General Reasoning}

\textbf{SuperGPQA \citep{du2025supergpqa}.} A question answering
benchmark at the graduate level, spanning over two hundred academic
disciplines.

\textbf{MMLU-Pro \citep{wang2024mmlupro}.} A more challenging
refinement of the MMLU benchmark with broader reasoning across
fourteen subject areas.

\textbf{BBEH \citep{kazemi2025bbeh}.} BIG-Bench Extra Hard is a
$23$-task benchmark that replaces each BIG-Bench Hard task with a
harder counterpart targeting related reasoning skills.

\textbf{GPQA-Diamond \citep{rein2024gpqa}.} The most challenging
subset of the GPQA benchmark, designed to be unsolvable through
simple retrieval.

\textbf{HumanEval \citep{chen2021humaneval}.} A Python code
generation benchmark with handwritten programming problems, each
accompanied by unit tests.

\textbf{AGIEval \citep{zhong2024agieval}.} A benchmark centered on
human examinations that evaluates foundation models on tasks drawn
from standardized tests.

\begin{table}[t]
\centering
\caption{Reasoning benchmarks used for evaluation.}
\label{tab:benchmarks}
\small
\setlength{\tabcolsep}{6pt}
\renewcommand{\arraystretch}{1.05}
\begin{tabular}{lll}
\toprule
\textbf{Benchmark} & \textbf{Domain} & \textbf{Metric} \\
\midrule
\multicolumn{3}{l}{\textit{Mathematical reasoning}} \\
AMC23            & High school competition math & mean@32 \\
MATH-500       & Mixed math topics            & pass@1 \\
GSM8K          & Grade school word problems   & pass@1 \\
Minerva        & STEM quantitative reasoning  & pass@1 \\
OlympiadBench  & Olympiad math                & pass@1 \\
AIME24         & Advanced high school math    & mean@32 \\
AIME25         & Advanced high school math    & mean@32 \\
\midrule
\multicolumn{3}{l}{\textit{General reasoning}} \\
SuperGPQA      & Graduate Q\&A                & pass@1 \\
MMLU-Pro       & Mixed domain understanding   & pass@1 \\
BBEH           & Hardest BIG-Bench subset     & pass@1 \\
GPQA-Diamond   & Google-proof Q\&A            & pass@1 \\
HumanEval      & Python code generation       & pass@1 \\
AGIEval        & Human examination tasks      & pass@1 \\
\bottomrule
\end{tabular}
\end{table}

\section{Anchor Mechanism Analysis}
\label{app:anchor_analysis}

This section provides a deeper analysis of the iterative anchor that
underlies AnchorLoop. We quantify the compute and memory overhead it
introduces, compare alternative refresh schedules, measure the
efficiency gain of caching anchor rollouts, and describe the failure
modes where the anchor stops contributing useful signal.

\subsection{Historical-Advantage Structure}
\label{app:hist_structure}

We analyze matched records from the first outer iteration for
Qwen3-4B-Base and MiMo-7B-Base. For each backbone, records
from each of five runs are ordered by Executor update step and divided
into early, middle, and late thirds before the stage counts are
aggregated. A candidate matched group is valid when the current and
anchor groups correspond to the same task and Curriculum batch, both
records are complete, and both groups have a resolved majority answer.
Invalid outputs, unresolved ties, and incomplete matches are excluded.

We classify the historical advantage from the discrete reward-indicator
patterns. Its centered vector is zero only when all entries are exactly
zero. A nonzero historical vector is collinear when the current vector
is also nonzero and the two vectors are related by a positive scale.
Negative proportionality and a nonzero historical vector paired with a
zero current vector are classified as non-collinear. Thus, no numerical
tolerance is used. The same-majority rate is reported separately and
can overlap with the zero category.

Table~\ref{tab:hist_structure} reports the resulting statistics. Each
run contributes 120 candidate groups per stage, giving 600 per stage
and backbone. These descriptive counts characterize the analyzed
records but do not establish that the historical term causes the
performance gains in the main results.

\begin{table*}[t]
\centering
\caption{Historical-advantage structure in matched records from five
runs per backbone. Percentages use valid groups as the denominator. The
final three columns partition the valid groups. The same-majority column
may overlap with them.}
\label{tab:hist_structure}
\small
\setlength{\tabcolsep}{5pt}
\begin{tabular}{lccccc}
\toprule
\textbf{Executor stage} & \shortstack{\textbf{Valid}/\\\textbf{candidate}} &
\shortstack{\textbf{Same}\\\textbf{majority}} &
\shortstack{\textbf{Zero historical}\\\textbf{term}} &
\shortstack{\textbf{Nonzero and}\\\textbf{collinear}} &
\shortstack{\textbf{Nonzero and}\\\textbf{non-collinear}} \\
\midrule
\multicolumn{6}{l}{\textbf{Qwen3-4B-Base}} \\
Early  & $581/600$ & $422\;(72.6\%)$ & $438\;(75.4\%)$ & $91\;(15.7\%)$ & $52\;(9.0\%)$ \\
Middle & $574/600$ & $355\;(61.8\%)$ & $379\;(66.0\%)$ & $111\;(19.3\%)$ & $84\;(14.6\%)$ \\
Late   & $586/600$ & $347\;(59.2\%)$ & $369\;(63.0\%)$ & $103\;(17.6\%)$ & $114\;(19.5\%)$ \\
\midrule
\multicolumn{6}{l}{\textbf{MiMo-7B-Base}} \\
Early  & $578/600$ & $414\;(71.6\%)$ & $430\;(74.4\%)$ & $96\;(16.6\%)$ & $52\;(9.0\%)$ \\
Middle & $569/600$ & $352\;(61.9\%)$ & $367\;(64.5\%)$ & $118\;(20.7\%)$ & $84\;(14.8\%)$ \\
Late   & $583/600$ & $340\;(58.3\%)$ & $361\;(61.9\%)$ & $106\;(18.2\%)$ & $116\;(19.9\%)$ \\
\bottomrule
\end{tabular}
\end{table*}

\subsection{Cost Analysis}
\label{app:cost}

Table~\ref{tab:cost} compares AnchorLoop with Agent0 on
Qwen3-4B-Base using eight A100 80GB GPUs. Timings are five-run means
for comparable single-iteration training. The anchor adds one frozen
checkpoint and $G$ rollouts per task. Cached AnchorLoop takes $24.1$
hours per iteration versus $20.5$ for Agent0, a $17.6\%$ overhead.
The main performance tables evaluate both methods after the same three
outer iterations. Since their per-iteration costs differ, this
comparison does not match wall time or GPU hours. It supports the
reported performance comparison at three iterations, but does not
establish an advantage under a matched compute budget.

\begin{table}[h]
\centering
\caption{Compute and memory comparison between Agent0 and AnchorLoop
on Qwen3-4B-Base.}
\label{tab:cost}
\small
\setlength{\tabcolsep}{6pt}
\renewcommand{\arraystretch}{1.1}
\begin{tabular}{lcc}
\toprule
\textbf{Metric} & \textbf{Agent0} & \textbf{AnchorLoop (ours)} \\
\midrule
Time per outer iteration       & $20.5$ h     & $24.1$ h \\
GPU hours per outer iteration  & $164$        & $193$ \\
Frozen checkpoint memory       & $0$          & $8.1$ GB \\
Anchor rollouts per task       & $0$          & $G = 8$ (cached) \\
Overhead vs Agent0             & --           & $+17.6\%$ \\
\bottomrule
\end{tabular}
\end{table}

\subsection{Anchor Refresh Schedule}
\label{app:refresh}

The default AnchorLoop refreshes the anchor at the start of every outer
iteration. Table~\ref{tab:refresh_schedule} compares this default
against refreshing every two iterations and against never refreshing,
which reduces to a fixed anchor pinned to the base model. For each
refresh schedule, we report the highest Math AVG among checkpoints
evaluated within the first five outer iterations. The every-iteration
and fixed-anchor results use the same runs and evaluation settings as
Figure~\ref{fig:analysis}(a).
The default schedule has the highest reported score in this table.
The three discrete frequencies differ in when the reference checkpoint
is updated. Their scores do not isolate checkpoint age or establish an
optimal rule beyond the evaluated settings.

\begin{table}[t]
\centering
\caption{Highest Math AVG among checkpoints evaluated within the first five outer iterations on Qwen3-4B-Base for each anchor refresh schedule.}
\label{tab:refresh_schedule}
\small
\setlength{\tabcolsep}{8pt}
\renewcommand{\arraystretch}{1.1}
\begin{tabular}{lc}
\toprule
\textbf{Refresh schedule} & \textbf{Best Math AVG} \\
\midrule
Every iteration (default)         & $\mathbf{56.3}$ \\
Every two iterations              & $54.4$ \\
No refresh (fixed at base model)  & $53.3$ \\
\bottomrule
\end{tabular}
\end{table}

\subsection{Anchor Caching Efficiency}
\label{app:caching}

Table~\ref{tab:caching} compares cached and uncached AnchorLoop on
the same eight GPUs and comparable single-iteration training. The
cached version samples $G=8$ anchor rollouts once per task and
Curriculum batch, then reuses them across relevant Executor steps. The
uncached version resamples $G=8$ at each such step. The five-run mean
falls from $32.7$ to $24.1$ hours, a $26.3\%$ reduction. Since
resampling changes stochastic outputs, this comparison of complete
implementations neither isolates a pure systems effect nor establishes
a theoretical lower bound on anchor overhead.

\begin{table}[t]
\centering
\caption{Wall time with and without anchor rollout caching on
Qwen3-4B-Base, averaged over five runs.}
\label{tab:caching}
\small
\setlength{\tabcolsep}{6pt}
\renewcommand{\arraystretch}{1.1}
\begin{tabular}{lc}
\toprule
\textbf{Configuration} & \shortstack{\textbf{Wall time per}\\\textbf{outer iteration (h)}} \\
\midrule
AnchorLoop without anchor caching & $32.7$ \\
AnchorLoop with anchor caching     & $\mathbf{24.1}$ \\
\midrule
Agent0 (reference)                & $20.5$ \\
\bottomrule
\end{tabular}
\end{table}

\subsection{Failure Case Analysis}
\label{app:failure}

The iterative anchor stops contributing useful signal in two
identifiable regimes. We describe each with a representative example
taken from our Curriculum logs on Qwen3-4B-Base.

\begin{tcolorbox}[breakable, colback=red!3, colframe=red!30,
  title=\textbf{Failure Case 1: Beyond the Frontier},
  title after break=\textbf{Failure Case 1: Beyond the Frontier (continued)},
  fonttitle=\bfseries, fontupper=\small]
\textit{Task.} A problem at the level of AIME 2024 Problem 15, which
asks for a unique point on a line segment that is not contained in
any unit segment with endpoints on the coordinate axes. The expected
solution combines coordinate geometry with optimisation.\\[4pt]
\textit{Anchor} ($\pi^{(t-1)}$). The eight rollouts produce eight
different answers, so the group has no unique modal label.\\[2pt]
\textit{Current} ($\pi^{(t)}$). The sampled answers also tie at the
highest count, leaving this group unresolved.\\[4pt]
Under the protocol in Appendix~\ref{app:answer_protocol}, these groups
are not assigned pseudo-labels. The task receives
$R_{\mathrm{front}}=0$, while the other Curriculum reward terms remain
active, and it is omitted from the pseudo-label-dependent Executor
update. The agreement-based terms therefore provide no comparison when
the sampled groups do not produce unique modal labels.
\end{tcolorbox}

\begin{tcolorbox}[breakable, colback=red!3, colframe=red!30,
  title=\textbf{Failure Case 2: Shared Wrong Consensus},
  fonttitle=\bfseries, fontupper=\small]
\textit{Task.} How many ways are there to seat $5$ distinct people
around a round table where rotations are considered equivalent? The
correct answer is $4! = 24$, after dividing by the size of the
rotation group.\\[4pt]
\textit{Anchor.} Most rollouts output $120$ and treat the table as
linear.\\[2pt]
\textit{Current.} Most rollouts also output $120$. Both checkpoints
inherit the same bias from the base model and forget to divide by the
size of the rotation group.\\[4pt]
The two groups therefore select the same wrong modal label. The two
indicators on every current trajectory coincide, which makes
$A^{\mathrm{hist}}=0$. The frontier term can compare how concentrated
the two sampled groups are, but it cannot determine that their shared
label is wrong. This case exposes a limitation of the present
agreement-based supervision. An anchor with the same bias cannot supply
the missing correctness signal.
\end{tcolorbox}

These cases illustrate two distinct sources of uninformative anchor
feedback. In the first case, the geometry and optimization task
yields dispersed, tied answers, leaving neither sampled group with
a unique modal label. In the second case, both groups count linear
arrangements instead of circular arrangements and overlook rotational
equivalence. Their shared incorrect modal answer makes
$A^\mathrm{hist}=0$, while $R_\mathrm{ref}$ can still compare agreement
rates without identifying the mistake. These examples illustrate
that useful anchor feedback depends on the sampled answer composition
and that shared reasoning errors can remain undetected.

\section{Additional Experimental Results}
\label{app:additional}

This section reports results that complement the main experiments in
Section~\ref{sec:experiments}. Section~\ref{app:additional:perbench} tracks
each individual mathematical benchmark across the three outer
iterations on MiMo-7B-Base, showing that the average improvements in
Table~\ref{tab:main_results} are not concentrated on one or two
benchmarks. Section~\ref{app:additional:dynamics_mimo} extends the
three iteration-dynamics panels of Figure~\ref{fig:dynamics} from
Qwen3-4B-Base to MiMo-7B-Base. Section~\ref{app:additional:sens_mimo}
extends the hyperparameter sensitivity heatmap of
Figure~\ref{fig:analysis}(b) to MiMo-7B-Base. All results in this
section use the same training protocol, anchor refresh schedule, and
evaluation conventions as Section~\ref{sec:exp:setup}.

\begin{table*}[t]
\centering
\caption{Per-benchmark math accuracy across outer iterations on
MiMo-7B-Base. Iteration $t=0$ is the Base Model with tool. Iteration
$t=3$ matches the endpoints in Table~\ref{tab:main_results}. Best
endpoint per benchmark is in \textbf{bold}.}
\label{tab:perbench_mimo}
\setlength{\tabcolsep}{3pt}
\renewcommand{\arraystretch}{1.05}
\small
\begin{tabular}{l c c c c c c c c}
\toprule
\textbf{Method (Iter)} & \textbf{AMC23} & \textbf{MATH} & \textbf{GSM8K} & \textbf{Minerva} & \textbf{Olympiad} & \textbf{AIME24} & \textbf{AIME25} & \textbf{AVG} \\
\midrule
\multicolumn{9}{l}{\textit{Agent0}} \\
\quad $t = 0$ & 63.6 & 77.9 & 91.1 & 51.4 & 50.2 & 34.4 & 26.9 & 56.5 \\
\quad $t = 1$ & 66.2 & 80.3 & 92.9 & 55.4 & 52.6 & 38.2 & 30.2 & 59.4 \\
\quad $t = 2$ & 67.0 & 81.0 & 93.8 & 56.9 & 53.8 & 38.9 & 32.0 & 60.5 \\
\quad $t = 3$ & 67.3 & 81.4 & 94.2 & 57.9 & 54.1 & 39.5 & 32.4 & 61.0 \\
\midrule
\multicolumn{9}{l}{\textit{AnchorLoop (ours)}} \\
\quad $t = 0$ & 63.6 & 77.9 & 91.1 & 51.4 & 50.2 & 34.4 & 26.9 & 56.5 \\
\quad $t = 1$ & 66.9 & 81.0 & 92.9 & 55.4 & 52.6 & 38.4 & 31.4 & 59.8 \\
\quad $t = 2$ & 68.8 & 82.8 & 94.1 & 58.1 & 54.9 & 40.7 & 33.9 & 61.9 \\
\rowcolor{verirow}
\quad $t = 3$ & \textbf{70.0} & \textbf{84.1} & \textbf{95.7} & \textbf{60.6} & \textbf{56.7} & \textbf{42.3} & \textbf{35.4} & \textbf{63.5} \\
\bottomrule
\end{tabular}
\end{table*}

\subsection{Per-Benchmark Trajectories on MiMo-7B-Base}
\label{app:additional:perbench}

Table~\ref{tab:perbench_mimo} reports math accuracy for each of the
seven mathematical reasoning benchmarks at outer iterations $t=0,1,2,3$
under both Agent0 and AnchorLoop on MiMo-7B-Base. Iteration $t=0$ is the
tool-augmented base model used as the shared starting point of both
methods. Iteration $t=3$ matches the endpoints in
Table~\ref{tab:main_results}.

Three observations stand out. First, every benchmark improves
monotonically across iterations under both methods, so the iteration
average gains in Figure~\ref{fig:dynamics_mimo}(a) are not driven by
trading off one column against another. Second, AnchorLoop's per-step gain is consistently larger than
Agent0's after iteration~$1$. The average gap widens from $0.4$ at $t=1$ to
$1.4$ at $t=2$ and $2.5$ at $t=3$, which matches the increasing
separation reported in Figure~\ref{fig:dynamics_mimo}(a).
Third, the largest absolute gains over the tool-augmented base appear
on the harder benchmarks. AnchorLoop lifts Minerva by $9.2$ points,
AIME25 by $8.5$, and AIME24 by $7.9$, while the easier GSM8K column
moves by only $4.6$ points because it already starts above $91\%$.
Agent0 also gains the most on these harder benchmarks, but consistently
by smaller margins, so the AnchorLoop advantage is largest precisely on
the columns with the most headroom.

\subsection{Iteration Dynamics on MiMo-7B-Base}
\label{app:additional:dynamics_mimo}

\begin{figure*}[t]
\centering
\AnchorLoopDynamicsFigure{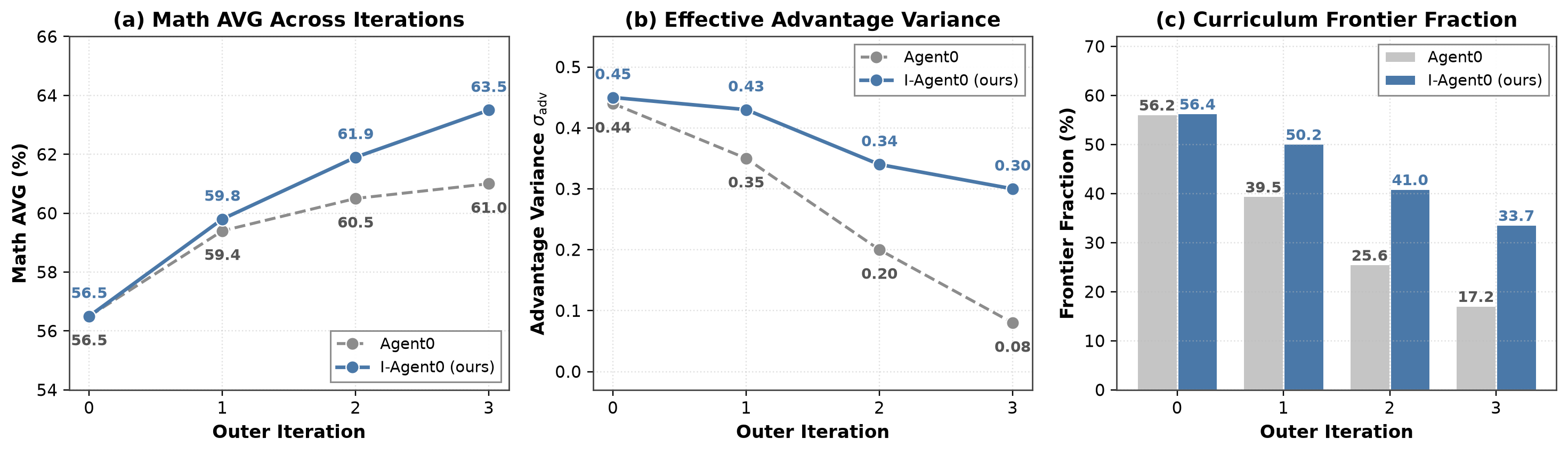}
\caption{Iteration dynamics on MiMo-7B-Base. (a) Math AVG across the three outer
iterations. (b) The within-group variance of effective advantages,
$V_{\mathrm{adv}}$, averaged over sampled groups in the Executor stage.
(c) The reported Curriculum frontier fraction, defined as the fraction
of generated tasks with a positive sampled agreement gap under
Appendix~\ref{app:answer_protocol}. An unresolved pair contributes zero.}
\label{fig:dynamics_mimo}
\Description{Three panels compare Agent0 and AnchorLoop on MiMo-7B-Base across outer iterations using math average, effective advantage variance, and the fraction of tasks with a positive sampled agreement gap.}
\end{figure*}

\paragraph{Iteration-wise math performance.}
Figure~\ref{fig:dynamics_mimo}(a) tracks math AVG on MiMo-7B-Base.
Both methods start from the tool-augmented base at $56.5$. Agent0
improves by $+2.9$ on iteration~$1$, then by only $+1.1$ and $+0.5$,
the same plateau pattern observed on Qwen3-4B-Base. AnchorLoop improves
by $+3.3$ points on iteration~$1$ and continues to gain $+2.1$ and
$+1.6$ points in the next two iterations. The gap to Agent0 therefore
widens from $0$ at iteration $0$ to $2.5$ at iteration $3$. The
iteration-$3$ endpoints match
Table~\ref{tab:main_results}, and the same gap of $2.5$ points carries
over from Qwen3-4B-Base, indicating that the iterative anchor scales
with backbone capability rather than producing a fixed offset.

\paragraph{Effective advantage variance.}
Figure~\ref{fig:dynamics_mimo}(b) reports the same stage-averaged
within-group variance $V_{\mathrm{adv}}$ defined in
Section~\ref{sec:exp:dynamics}. Agent0 begins at $0.44$, while
AnchorLoop begins at $0.45$. By iteration~$3$, Agent0 decreases to
$0.08$, whereas AnchorLoop remains at $0.30$. These stage averages
summarize the sampled Executor groups and
do not imply that the historical term is nonzero in every group. Under
the definitions in Section~\ref{sec:method:executor}, both advantage
terms vanish when all current answers in a group are identical. The
observed difference is consistent with greater within-group variation
for AnchorLoop in the evaluated MiMo runs. This panel alone does not
isolate a causal effect on performance.

\paragraph{Curriculum frontier fraction.}
Figure~\ref{fig:dynamics_mimo}(c) follows the construction in
Figure~\ref{fig:dynamics}(c). Its two groups are sampled independently
from the same checkpoint, and an unresolved pair contributes zero.
At iteration~$0$, the reported fraction is $56.2\%$ for Agent0 and
$56.4\%$ for AnchorLoop. By iteration~$3$, it decreases to $17.2\%$
for Agent0 and to $33.7\%$ for AnchorLoop. These values record
how often independently sampled groups produce a positive majority
agreement gap. They do not measure correctness, identify newly solved
tasks, or establish that the frontier reward caused the later
performance difference. At iteration~3 on both backbones, AnchorLoop reports higher math AVG,
higher effective advantage variance, and a larger positive agreement
gap fraction. These concurrent differences do not establish a causal
relation among the statistics.

\subsection{Hyperparameter Sensitivity on MiMo-7B-Base}
\label{app:additional:sens_mimo}

\begin{figure}[h]
\centering
\includegraphics[width=0.9\linewidth]{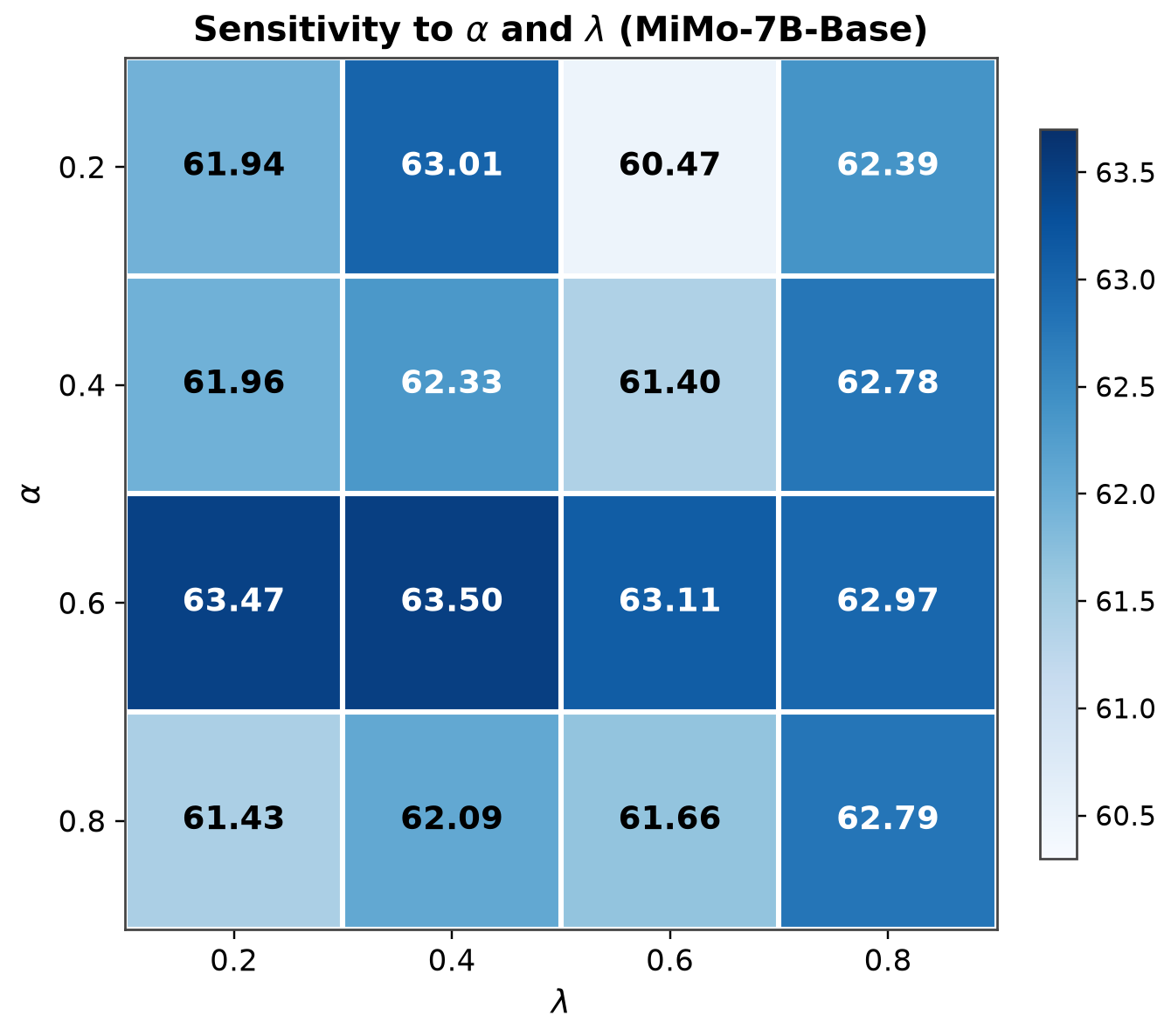}
\caption{Hyperparameter sensitivity on MiMo-7B-Base, the appendix
counterpart of Figure~\ref{fig:analysis}(b). Math AVG on a $4\times 4$
grid over the historical-advantage weight $\alpha$ and the frontier reward
weight $\lambda$. The peak at our default $(\alpha,\lambda) = (0.6, 0.4)$
reaches $63.50$, matching Table~\ref{tab:main_results}.}
\label{fig:sens_mimo}
\Description{A four-by-four heatmap shows math average on MiMo-7B-Base over combinations of the historical-advantage weight alpha and frontier reward weight lambda.}
\end{figure}

We sweep $\alpha \in \{0.2, 0.4, 0.6, 0.8\}$ and
$\lambda \in \{0.2, 0.4, 0.6, 0.8\}$ on MiMo-7B-Base, the same grid
used for Qwen3-4B-Base in Figure~\ref{fig:analysis}(b).
Figure~\ref{fig:sens_mimo} reports math AVG at each grid cell. The
default setting $(\alpha, \lambda) = (0.6, 0.4)$ attains the highest
reported Math AVG on both evaluated grids. On MiMo-7B-Base, Math AVG
ranges from $60.47$ to $63.50$, indicating greater sensitivity than
on Qwen3-4B-Base. The default setting remains the highest-performing
point on both evaluated grids. These results do not establish
robustness outside the tested ranges or prove that the weights transfer
across backbones without selection.

\section{Case Studies and Qualitative Examples}
\label{app:case}

\definecolor{anchorbg}{HTML}{ECECEC}
\definecolor{anchoredge}{HTML}{8C8C8C}
\definecolor{currentbg}{HTML}{DCE9F4}
\definecolor{currentedge}{HTML}{4A78A8}
\definecolor{taskbg}{HTML}{FFF9E6}
\definecolor{taskedge}{HTML}{C9A227}
\definecolor{pythonbg}{HTML}{FFF5DC}
\definecolor{pythonedge}{HTML}{B89B5E}
\definecolor{curribg}{HTML}{DCEFDC}
\definecolor{curriedge}{HTML}{5A8C5A}

\tcbset{
  frontiercase/.style={
    enhanced,
    colback=white,
    colframe=black!55,
    boxrule=0.4pt,
    arc=2pt,
    left=4pt,right=4pt,top=4pt,bottom=4pt,
    fonttitle=\bfseries\small,
    breakable,
    before skip=6pt,after skip=6pt
  },
  taskstrip/.style={
    enhanced,
    colback=taskbg,
    colframe=taskedge,
    boxrule=0.4pt,
    arc=1pt,
    left=4pt,right=4pt,top=2pt,bottom=2pt,
    fontupper=\small
  },
  anchorstrip/.style={
    enhanced,
    colback=anchorbg,
    colframe=anchoredge,
    boxrule=0.4pt,
    arc=1pt,
    left=4pt,right=4pt,top=2pt,bottom=2pt,
    fontupper=\small
  },
  currentstrip/.style={
    enhanced,
    colback=currentbg,
    colframe=currentedge,
    boxrule=0.4pt,
    arc=1pt,
    left=4pt,right=4pt,top=2pt,bottom=2pt,
    fontupper=\small
  },
  pythonstrip/.style={
    enhanced,
    colback=pythonbg,
    colframe=pythonedge,
    boxrule=0.4pt,
    arc=1pt,
    left=4pt,right=4pt,top=2pt,bottom=2pt,
    fontupper=\small\ttfamily
  },
  curristrip/.style={
    enhanced,
    colback=curribg,
    colframe=curriedge,
    boxrule=0.4pt,
    arc=1pt,
    left=4pt,right=4pt,top=2pt,bottom=2pt,
    fontupper=\small
  }
}

This appendix provides case-level context for the quantitative analyses
in Sections~\ref{sec:exp:dynamics} and~\ref{sec:exp:sensitivity}.
Section~\ref{app:case:frontier} compares
twelve Curriculum tasks across three outer iterations and several
mathematical topics. Section~\ref{app:case:evolution} groups eight topic
sequences, each containing one task from iterations $0$ through $3$.
The grouping organizes related prompts for comparison and does not
establish that a later task was produced by editing the earlier tasks.
Section~\ref{app:case:traces} reports four full reasoning traces, with
three AIME 2024 problems and one Curriculum task from iteration $3$.
It places selected anchor and post-update Executor traces side by side.
All cases are drawn from runs on Qwen3-4B-Base.

\subsection{Extended Post-update Executor Comparisons}
\label{app:case:frontier}

We report twelve selected Curriculum tasks, with four examples from
each outer iteration $t \in \{1, 2, 3\}$. For each task, we show a
displayed anchor output in gray and a displayed post-update Executor
output in blue. The visual code matches the gray and blue robots used
in Figure~\ref{fig:framework}. These
individual outputs illustrate case-level differences. They do not by
themselves verify the group-level condition
$\mu^{\mathrm{cur}}(x) > \mu^{\mathrm{anc}}(x)$, estimate how often the
differences occur, or establish that $R_{\mathrm{ref}}$ caused them.

\paragraph{Iteration~$1$ (anchor $=$ base Executor, current $=$ iteration-$1$ Executor).}
The selected iteration-$1$ cases display gaps such as a missing
inclusion-exclusion correction or leading-digit constraint. The paired
updated Executor outputs include the corresponding structural steps.

\begin{tcolorbox}[frontiercase,title={Selected Example~1 (Iter 1) -- Inclusion-Exclusion}]
\begin{tcolorbox}[taskstrip]
\textbf{Task.} Find the sum of all integers from $1$ to $60$ that are
divisible by $4$ or by $6$.
\end{tcolorbox}
\vspace{2pt}
\begin{tcolorbox}[anchorstrip]
\textbf{Anchor $\pi^{(0)}$.}
Multiples of $4$ up to $60$: $4+8+\dots+60 = 4\cdot\tfrac{15\cdot 16}{2} = 480$.
Multiples of $6$ up to $60$: $6+12+\dots+60 = 6\cdot\tfrac{10\cdot 11}{2} = 330$.
Sum $= 480 + 330 = \boxed{810}$. \textcolor{red}{\textbf{Wrong}}: forgot
to subtract multiples of $\mathrm{lcm}(4,6)=12$.
\end{tcolorbox}
\vspace{2pt}
\begin{tcolorbox}[currentstrip]
\textbf{Current $\pi^{(1)}$.}
Multiples of $4$: $480$. Multiples of $6$: $330$. Multiples of
$\mathrm{lcm}(4,6)=12$: $12+24+\dots+60 = 12\cdot\tfrac{5\cdot 6}{2}=180$.
Inclusion-exclusion: $480 + 330 - 180 = \boxed{630}$. \textcolor{currentedge!80!black}{\textbf{Correct}}.
\end{tcolorbox}
\end{tcolorbox}

\begin{tcolorbox}[frontiercase,title={Selected Example~2 (Iter 1) -- Arithmetic Sequence Off-by-One}]
\begin{tcolorbox}[taskstrip]
\textbf{Task.} An arithmetic progression has first term $a_1 = 7$ and
common difference $d = 3$. Find the $50$th term $a_{50}$.
\end{tcolorbox}
\vspace{2pt}
\begin{tcolorbox}[anchorstrip]
\textbf{Anchor $\pi^{(0)}$.}
$a_n = a_1 + n \cdot d = 7 + 50 \cdot 3 = \boxed{157}$.
\textcolor{red}{\textbf{Wrong}}: used $n \cdot d$ instead of $(n-1)\cdot d$.
\end{tcolorbox}
\vspace{2pt}
\begin{tcolorbox}[currentstrip]
\textbf{Current $\pi^{(1)}$.}
$a_n = a_1 + (n-1)\cdot d$. So $a_{50} = 7 + 49 \cdot 3 = 7 + 147 = \boxed{154}$.
\textcolor{currentedge!80!black}{\textbf{Correct}}.
\end{tcolorbox}
\end{tcolorbox}

\begin{tcolorbox}[frontiercase,title={Selected Example~3 (Iter 1) -- Restricted Digit Count}]
\begin{tcolorbox}[taskstrip]
\textbf{Task.} How many $5$-digit positive integers have all distinct
digits? (Leading digit cannot be $0$.)
\end{tcolorbox}
\vspace{2pt}
\begin{tcolorbox}[anchorstrip]
\textbf{Anchor $\pi^{(0)}$.}
Choose $5$ distinct digits and order them: $10 \cdot 9 \cdot 8 \cdot
7 \cdot 6 = \boxed{30{,}240}$. \textcolor{red}{\textbf{Wrong}}: counts
numbers starting with $0$, which are not $5$-digit.
\end{tcolorbox}
\vspace{2pt}
\begin{tcolorbox}[currentstrip]
\textbf{Current $\pi^{(1)}$.}
First digit $\in \{1,\dots,9\}$: $9$ choices. Remaining four positions
take any of the remaining nine digits without repeats:
$9 \cdot 9 \cdot 8 \cdot 7 \cdot 6 = \boxed{27{,}216}$.
\textcolor{currentedge!80!black}{\textbf{Correct}}.
\end{tcolorbox}
\end{tcolorbox}

\begin{tcolorbox}[frontiercase,title={Selected Example~4 (Iter 1) -- Modular Pattern}]
\begin{tcolorbox}[taskstrip]
\textbf{Task.} How many integers $n$ with $1 \le n \le 20$ satisfy
$n^2 \equiv 1 \pmod{8}$?
\end{tcolorbox}
\vspace{2pt}
\begin{tcolorbox}[anchorstrip]
\textbf{Anchor $\pi^{(0)}$.}
Try the first few squares: $1^2 = 1$, $3^2 = 9 \equiv 1$, $9^2 = 81
\equiv 1$. Only $n \in \{1, 3, 9\}$ matter, so the answer is $\boxed{3}$.
\textcolor{red}{\textbf{Wrong}}: lists ad hoc solutions instead of
identifying the structural rule.
\end{tcolorbox}
\vspace{2pt}
\begin{tcolorbox}[currentstrip]
\textbf{Current $\pi^{(1)}$.}
Every odd $n$ satisfies $n^2 \equiv 1 \pmod 8$, since
$(2k+1)^2 = 4k(k+1) + 1$ and $k(k+1)$ is even. Even $n$ have
$n^2 \equiv 0$ or $4 \pmod 8$. So count odd integers in $[1, 20]$:
$\{1, 3, 5, \dots, 19\}$, which gives $\boxed{10}$.
\textcolor{currentedge!80!black}{\textbf{Correct}}.
\end{tcolorbox}
\end{tcolorbox}

\paragraph{Iteration~$2$ (anchor $=$ iteration-$1$ Executor, current $=$ iteration-$2$ Executor).}
The selected iteration-$2$ cases display errors involving cyclic
constraints, auxiliary identities, and complementary counting. The
paired updated Executor outputs use the corresponding additional
steps.

\begin{tcolorbox}[frontiercase,title={Selected Example~5 (Iter 2) -- Cyclic Bead Coloring}]
\begin{tcolorbox}[taskstrip]
\textbf{Task.} How many ways are there to color the $5$ beads of a
necklace arranged in a cycle, using $3$ colors, so that no two
adjacent beads have the same color? (Beads are distinguishable by
position; rotations are not identified.)
\end{tcolorbox}
\vspace{2pt}
\begin{tcolorbox}[anchorstrip]
\textbf{Anchor $\pi^{(1)}$.}
First bead: $3$ choices. Each of the next four: $2$ choices (different
from the previous). Total: $3 \cdot 2^4 = \boxed{48}$.
\textcolor{red}{\textbf{Wrong}}: treats the necklace as a line and
ignores the constraint between bead~$5$ and bead~$1$.
\end{tcolorbox}
\vspace{2pt}
\begin{tcolorbox}[currentstrip]
\textbf{Current $\pi^{(2)}$.}
Chromatic polynomial of cycle $C_n$ with $k$ colors is $P(C_n,k) =
(k-1)^n + (-1)^n (k-1)$. For $n = 5$, $k = 3$: $2^5 + (-1)^5 \cdot 2 =
32 - 2 = \boxed{30}$. \textcolor{currentedge!80!black}{\textbf{Correct}}.
\end{tcolorbox}
\end{tcolorbox}

\begin{tcolorbox}[frontiercase,title={Selected Example~6 (Iter 2) -- Telescoping Sum}]
\begin{tcolorbox}[taskstrip]
\textbf{Task.} Compute $\displaystyle\sum_{n=1}^{20} \frac{1}{n(n+1)}$
exactly as a reduced fraction.
\end{tcolorbox}
\vspace{2pt}
\begin{tcolorbox}[anchorstrip]
\textbf{Anchor $\pi^{(1)}$.}
Add the first few decimals: $0.5 + 0.1667 + 0.0833 + \dots \approx 0.95$.
Final answer: $\boxed{0.95}$.
\textcolor{red}{\textbf{Wrong}}: returns a decimal approximation and
never uses the telescoping identity.
\end{tcolorbox}
\vspace{2pt}
\begin{tcolorbox}[currentstrip]
\textbf{Current $\pi^{(2)}$.}
Partial fractions: $\tfrac{1}{n(n+1)} = \tfrac{1}{n} - \tfrac{1}{n+1}$.
The sum telescopes: $\sum_{n=1}^{20} \big(\tfrac{1}{n} - \tfrac{1}{n+1}\big)
= 1 - \tfrac{1}{21} = \boxed{\tfrac{20}{21}}$.
\textcolor{currentedge!80!black}{\textbf{Correct}}.
\end{tcolorbox}
\end{tcolorbox}

\begin{tcolorbox}[frontiercase,title={Selected Example~7 (Iter 2) -- Constrained Permutation}]
\begin{tcolorbox}[taskstrip]
\textbf{Task.} How many distinct arrangements of the letters in
\textsc{committee} have the two \textsc{M}s separated by at least one
other letter?
\end{tcolorbox}
\vspace{2pt}
\begin{tcolorbox}[anchorstrip]
\textbf{Anchor $\pi^{(1)}$.}
The word has $9$ letters with repeats $\{M\!:\!2, T\!:\!2, E\!:\!2\}$.
Total arrangements: $9!/(2!\,2!\,2!) = 362880/8 = \boxed{45{,}360}$.
\textcolor{red}{\textbf{Wrong}}: ignores the
``\textsc{M}s separated'' constraint entirely.
\end{tcolorbox}
\vspace{2pt}
\begin{tcolorbox}[currentstrip]
\textbf{Current $\pi^{(2)}$.}
Use complement. Total $= 45360$. Arrangements with the two \textsc{M}s
adjacent: treat \textsc{MM} as one block, leaving $8$ items with repeats
$\{T\!:\!2, E\!:\!2\}$: $8!/(2!\,2!) = 40320/4 = 10080$. Answer:
$45360 - 10080 = \boxed{35{,}280}$.
\textcolor{currentedge!80!black}{\textbf{Correct}}.
\end{tcolorbox}
\end{tcolorbox}

\begin{tcolorbox}[frontiercase,title={Selected Example~8 (Iter 2) -- Modular Exponentiation}]
\begin{tcolorbox}[taskstrip]
\textbf{Task.} Compute $7^{100} \bmod 13$.
\end{tcolorbox}
\vspace{2pt}
\begin{tcolorbox}[anchorstrip]
\textbf{Anchor $\pi^{(1)}$.}
Multiply step by step without Fermat. $7^2 \equiv 10$,
$7^3 \equiv 7 \cdot 10 = 70 \equiv 5$, $7^4 \equiv 35 \equiv 9$,
$7^5 \equiv 63 \equiv 11$, and so on for $100$ steps. The running
remainder drifts off course around step $40$; the anchor's final
value is $\boxed{4}$.
\textcolor{red}{\textbf{Wrong}}: a brute hand iteration without the
period-$12$ structure accumulates an arithmetic slip after dozens of
multiplications.
\end{tcolorbox}
\vspace{2pt}
\begin{tcolorbox}[currentstrip]
\textbf{Current $\pi^{(2)}$.}
By Fermat's little theorem, $7^{12} \equiv 1 \pmod{13}$. Write
$100 = 8\cdot 12 + 4$, so $7^{100} \equiv 7^4 \pmod{13}$.
$7^2 = 49 \equiv 10$, $7^4 \equiv 100 \equiv 9 \pmod{13}$. Answer
$\boxed{9}$. \textcolor{currentedge!80!black}{\textbf{Correct}}.
\end{tcolorbox}
\end{tcolorbox}

\paragraph{Iteration~$3$ (anchor $=$ iteration-$2$ Executor, current $=$ iteration-$3$ Executor).}
The selected iteration-$3$ cases emphasize verification and broader
search. Their updated Executor outputs use the sandbox to check
intermediate claims or combine a symbolic route with a numerical
check. This observation applies only to the displayed cases.

\begin{tcolorbox}[frontiercase,title={Selected Example~9 (Iter 3) -- Sum of Trailing-Zero Counts}]
\begin{tcolorbox}[taskstrip]
\textbf{Task.} Let $z(n)$ denote the number of trailing zeros of $n!$
in base $10$. Compute $\sum_{n=1}^{100} z(n)$.
\end{tcolorbox}
\vspace{2pt}
\begin{tcolorbox}[anchorstrip]
\textbf{Anchor $\pi^{(2)}$.}
Approximate $z(n) \approx n/4$ over the range, so the sum is about
$\tfrac{100\cdot 101}{2}/4 \approx 1263$. Final $\boxed{1263}$.
\textcolor{red}{\textbf{Wrong}}: estimates the average rate of zeros
and never sums Legendre's exact formula.
\end{tcolorbox}
\vspace{2pt}
\begin{tcolorbox}[currentstrip]
\textbf{Current $\pi^{(3)}$.}
Legendre: $z(n) = \lfloor n/5\rfloor + \lfloor n/25\rfloor + \lfloor n/125\rfloor$.
Verify with Python:\\[2pt]
\begin{tcolorbox}[pythonstrip]
def z(n):\\
\hspace*{1em}t=0; p=5\\
\hspace*{1em}while p<=n: t+=n//p; p*=5\\
\hspace*{1em}return t\\
print(sum(z(n) for n in range(1,101)))  \# 1124
\end{tcolorbox}
Answer $\boxed{1124}$. \textcolor{currentedge!80!black}{\textbf{Correct}}.
\end{tcolorbox}
\end{tcolorbox}

\begin{tcolorbox}[frontiercase,title={Selected Example~10 (Iter 3) -- Palindromic Quartic}]
\begin{tcolorbox}[taskstrip]
\textbf{Task.} Compute the sum of all real roots of
$x^4 - 7x^3 + 14x^2 - 7x + 1 = 0$.
\end{tcolorbox}
\vspace{2pt}
\begin{tcolorbox}[anchorstrip]
\textbf{Anchor $\pi^{(2)}$.}
Try rational roots $\pm 1$. $p(1) = 1-7+14-7+1=2$, $p(-1)=1+7+14+7+1=30$.
No rational roots, so default to Vieta on the full polynomial: sum of
\emph{all} roots is $\boxed{7}$.
\textcolor{red}{\textbf{Partially correct, but unjustified.}} The
answer happens to be right because the polynomial has four real roots,
but the anchor never verifies reality.
\end{tcolorbox}
\vspace{2pt}
\begin{tcolorbox}[currentstrip]
\textbf{Current $\pi^{(3)}$.}
The polynomial is palindromic. Divide by $x^2$ and substitute $y = x +
1/x$: $y^2 - 7y + 12 = 0$, giving $y \in \{3, 4\}$.
$x + 1/x = 3 \Rightarrow x^2 - 3x + 1 = 0$, roots $(3 \pm \sqrt{5})/2$,
both real. $x + 1/x = 4 \Rightarrow x^2 - 4x + 1 = 0$, roots
$2 \pm \sqrt{3}$, both real. Sum: $3 + 4 = \boxed{7}$. Verify with
Python:\\[2pt]
\begin{tcolorbox}[pythonstrip]
import numpy as np\\
print(sum(np.roots([1,-7,14,-7,1]).real))  \# 7.0
\end{tcolorbox}
\textcolor{currentedge!80!black}{\textbf{Correct, with reality
justified.}}
\end{tcolorbox}
\end{tcolorbox}

\begin{tcolorbox}[frontiercase,title={Selected Example~11 (Iter 3) -- Prime-Restricted Compositions}]
\begin{tcolorbox}[taskstrip]
\textbf{Task.} How many ordered $5$-tuples $(p_1, p_2, p_3, p_4, p_5)$
with each $p_i \in \{2, 3, 5, 7, 11\}$ satisfy
$p_1 + p_2 + p_3 + p_4 + p_5 = 30$?
\end{tcolorbox}
\vspace{2pt}
\begin{tcolorbox}[anchorstrip]
\textbf{Anchor $\pi^{(2)}$.}
Try the symmetric tuple $(6,6,6,6,6)$ -- but $6$ is not prime. Try
$(7,7,5,5,6)$, also bad. Enumerate by hand a few candidates such as
$(11,7,5,5,2)$ ($30$ \ding{51}) and $(7,7,7,7,2)$ ($30$ \ding{51});
each has multiple orderings. Estimate $\boxed{120}$.
\textcolor{red}{\textbf{Wrong}}: undercounts because the manual
enumeration misses several multisets.
\end{tcolorbox}
\vspace{2pt}
\begin{tcolorbox}[currentstrip]
\textbf{Current $\pi^{(3)}$.}
This is the coefficient of $x^{30}$ in $(x^2 + x^3 + x^5 + x^7 + x^{11})^5$.
Compute via Python:\\[2pt]
\begin{tcolorbox}[pythonstrip]
from itertools import product\\
P=[2,3,5,7,11]\\
print(sum(1 for t in product(P,repeat=5) if sum(t)==30))  \# 155
\end{tcolorbox}
Answer $\boxed{155}$. \textcolor{currentedge!80!black}{\textbf{Correct}}.
\end{tcolorbox}
\end{tcolorbox}

\begin{tcolorbox}[frontiercase,title={Selected Example~12 (Iter 3) -- Sum-of-Two-Squares Diophantine}]
\begin{tcolorbox}[taskstrip]
\textbf{Task.} Find all positive integer pairs $(x, y)$ with $x \le y$
satisfying $x^2 + y^2 = 2025$.
\end{tcolorbox}
\vspace{2pt}
\begin{tcolorbox}[anchorstrip]
\textbf{Anchor $\pi^{(2)}$.}
$2025 = 45^2$. So $(0, 45)$ is one pair, but $x$ must be positive.
Try $x = 9$: $y^2 = 2025 - 81 = 1944$, not a square.
Try $x = 15$: $y^2 = 1800$, not a square. Conclude there are
\boxed{\text{no solutions}}.
\textcolor{red}{\textbf{Wrong}}: stops after two trials.
\end{tcolorbox}
\vspace{2pt}
\begin{tcolorbox}[currentstrip]
\textbf{Current $\pi^{(3)}$.}
Search $x$ in $[1, \lfloor\sqrt{2025/2}\rfloor] = [1, 31]$ exhaustively
with Python:\\[2pt]
\begin{tcolorbox}[pythonstrip]
N=2025\\
sols=[(x,int((N-x*x)**0.5)) for x in range(1,32)\\
\hspace*{1em}if (N-x*x) >= 0 and int((N-x*x)**0.5)**2 == N-x*x]\\
sols=[(x,y) for x,y in sols if x<=y]\\
print(sols)  \# [(27, 36)]
\end{tcolorbox}
Unique pair $\boxed{(27, 36)}$, since $27^2 + 36^2 = 729 + 1296 = 2025$.
\textcolor{currentedge!80!black}{\textbf{Correct}}.
\end{tcolorbox}
\end{tcolorbox}

\paragraph{Observations in the twelve selected examples.}
The displayed pairs include missing identities, omitted constraints,
and incomplete verification. Several iteration-$3$ updated Executor
outputs use Python verification, while the corresponding displayed
anchor outputs rely on approximations or limited manual trials. The
examples were selected for illustration. Their composition does not
estimate the prevalence of these behaviors, establish that
$R_{\mathrm{ref}}$ caused them, or show that tool use first appears
at iteration~$3$.

\subsection{Selected Curriculum Tasks Across Outer Iterations}
\label{app:case:evolution}

We group thirty-two selected Curriculum tasks into eight topic
sequences. Each sequence contains one task from iterations
$t = 0, 1, 2, 3$ under a common mathematical topic. This retrospective
organization highlights differences among the displayed prompts. It
does not establish a direct editing lineage, a general increase in
task difficulty, or a causal effect of the Curriculum reward.

\begin{tcolorbox}[frontiercase,title={Topic Sequence~A -- Inclusion-Exclusion}]
\begin{tcolorbox}[curristrip]
\textbf{Iter $0$:} Compute the sum of all multiples of $3$ between
$1$ and $100$.
\end{tcolorbox}
\vspace{2pt}
\begin{tcolorbox}[curristrip]
\textbf{Iter $1$:} Compute the sum of all integers between $1$ and
$100$ that are divisible by $3$ \emph{or} by $5$.
\end{tcolorbox}
\vspace{2pt}
\begin{tcolorbox}[curristrip]
\textbf{Iter $2$:} Compute the sum of all integers between $1$ and
$100$ that are divisible by at least one of $3$, $5$, or $7$.
\end{tcolorbox}
\vspace{2pt}
\begin{tcolorbox}[curristrip]
\textbf{Iter $3$:} Count the integers between $1$ and $1000$
that are coprime to $2 \cdot 3 \cdot 5 \cdot 7 = 210$, and report the
result.
\end{tcolorbox}
\end{tcolorbox}

\begin{tcolorbox}[frontiercase,title={Topic Sequence~B -- Combinatorics with Restrictions}]
\begin{tcolorbox}[curristrip]
\textbf{Iter $0$:} How many ways are there to arrange the letters
\textsc{abcde} in a row?
\end{tcolorbox}
\vspace{2pt}
\begin{tcolorbox}[curristrip]
\textbf{Iter $1$:} How many distinct arrangements does the word
\textsc{banana} have?
\end{tcolorbox}
\vspace{2pt}
\begin{tcolorbox}[curristrip]
\textbf{Iter $2$:} How many distinct arrangements of the letters of
\textsc{mississippi} have no two \textsc{i}s adjacent?
\end{tcolorbox}
\vspace{2pt}
\begin{tcolorbox}[curristrip]
\textbf{Iter $3$:} Eight people sit at a round table. Two specific
pairs, $(A, B)$ and $(C, D)$, refuse to sit next to each other. How
many seatings are there if rotations are equivalent? Verify with a
short Python enumeration.
\end{tcolorbox}
\end{tcolorbox}

\begin{tcolorbox}[frontiercase,title={Topic Sequence~C -- Sequences and Recurrences}]
\begin{tcolorbox}[curristrip]
\textbf{Iter $0$:} The arithmetic progression $a_1 = 2, d = 5$ has
what value at $a_{10}$?
\end{tcolorbox}
\vspace{2pt}
\begin{tcolorbox}[curristrip]
\textbf{Iter $1$:} Compute the sum of the first $50$ terms of the
arithmetic progression with $a_1 = 2, d = 5$.
\end{tcolorbox}
\vspace{2pt}
\begin{tcolorbox}[curristrip]
\textbf{Iter $2$:} Express $\sum_{n=1}^{N} \tfrac{1}{n(n+1)}$ in
closed form and evaluate at $N = 100$.
\end{tcolorbox}
\vspace{2pt}
\begin{tcolorbox}[curristrip]
\textbf{Iter $3$:} The sequence $a_n = 3 a_{n-1} - 2 a_{n-2}$ with
$a_0 = 1, a_1 = 4$ has a closed form. Derive it, evaluate $a_{30}$,
and verify with Python.
\end{tcolorbox}
\end{tcolorbox}

\begin{tcolorbox}[frontiercase,title={Topic Sequence~D -- Probability and Expectation}]
\begin{tcolorbox}[curristrip]
\textbf{Iter $0$:} A fair six-sided die is rolled. What is the
probability of rolling a $6$?
\end{tcolorbox}
\vspace{2pt}
\begin{tcolorbox}[curristrip]
\textbf{Iter $1$:} Two fair six-sided dice are rolled. What is the
probability that the sum equals $7$?
\end{tcolorbox}
\vspace{2pt}
\begin{tcolorbox}[curristrip]
\textbf{Iter $2$:} A fair coin is flipped five times. What is the
probability of getting at least two heads?
\end{tcolorbox}
\vspace{2pt}
\begin{tcolorbox}[curristrip]
\textbf{Iter $3$:} Ten fair six-sided dice are rolled. Compute the
expected number of distinct face values that appear, as a reduced
fraction, and verify with a short Monte Carlo run in Python.
\end{tcolorbox}
\end{tcolorbox}

\begin{tcolorbox}[frontiercase,title={Topic Sequence~E -- Number Theory}]
\begin{tcolorbox}[curristrip]
\textbf{Iter $0$:} Compute $\gcd(48, 60)$.
\end{tcolorbox}
\vspace{2pt}
\begin{tcolorbox}[curristrip]
\textbf{Iter $1$:} Find the smallest positive integer $n > 1$ with
$n^2 \equiv 1 \pmod{13}$.
\end{tcolorbox}
\vspace{2pt}
\begin{tcolorbox}[curristrip]
\textbf{Iter $2$:} Find the number of positive divisors of $1440 =
2^5 \cdot 3^2 \cdot 5$, and their sum.
\end{tcolorbox}
\vspace{2pt}
\begin{tcolorbox}[curristrip]
\textbf{Iter $3$:} Find the smallest positive integer $n$ with
$\varphi(n) = 48$, where $\varphi$ is Euler's totient function.
Verify with Python over $n \le 100$.
\end{tcolorbox}
\end{tcolorbox}

\begin{tcolorbox}[frontiercase,title={Topic Sequence~F -- Algebra and Polynomials}]
\begin{tcolorbox}[curristrip]
\textbf{Iter $0$:} Solve $2x + 7 = 19$ for $x$.
\end{tcolorbox}
\vspace{2pt}
\begin{tcolorbox}[curristrip]
\textbf{Iter $1$:} Solve $x^2 - 5x + 6 = 0$ for $x$.
\end{tcolorbox}
\vspace{2pt}
\begin{tcolorbox}[curristrip]
\textbf{Iter $2$:} The roots of $x^2 - 7x + 10 = 0$ are $r$ and $s$.
Compute $r^2 + s^2$ using Vieta's formulas, without solving for the
roots explicitly.
\end{tcolorbox}
\vspace{2pt}
\begin{tcolorbox}[curristrip]
\textbf{Iter $3$:} Find every real root of $x^4 - 6 x^2 + 8 = 0$ and
report the sum of the real roots. Then count complex roots. Verify
with Python's \texttt{numpy.roots}.
\end{tcolorbox}
\end{tcolorbox}

\begin{tcolorbox}[frontiercase,title={Topic Sequence~G -- Geometry}]
\begin{tcolorbox}[curristrip]
\textbf{Iter $0$:} Compute the area of the triangle with vertices
$(0, 0)$, $(4, 0)$, and $(0, 3)$.
\end{tcolorbox}
\vspace{2pt}
\begin{tcolorbox}[curristrip]
\textbf{Iter $1$:} Compute the area of a regular hexagon with side
length $5$.
\end{tcolorbox}
\vspace{2pt}
\begin{tcolorbox}[curristrip]
\textbf{Iter $2$:} A circular sector has radius $12$ and central angle
$60^{\circ}$. Compute the arc length and the area of the sector,
leaving answers in terms of $\pi$.
\end{tcolorbox}
\vspace{2pt}
\begin{tcolorbox}[curristrip]
\textbf{Iter $3$:} A frustum has base radii $R_1 = 6$ and $R_2 = 3$
and slant height $\ell = 5$. Compute the volume in closed form. Verify
the height computation with Python and confirm the answer.
\end{tcolorbox}
\end{tcolorbox}

\begin{tcolorbox}[frontiercase,title={Topic Sequence~H -- Calculus}]
\begin{tcolorbox}[curristrip]
\textbf{Iter $0$:} Differentiate $f(x) = x^3 - 2x + 1$ with respect
to $x$.
\end{tcolorbox}
\vspace{2pt}
\begin{tcolorbox}[curristrip]
\textbf{Iter $1$:} Evaluate $\int_{0}^{4} (2x + 3)\, dx$.
\end{tcolorbox}
\vspace{2pt}
\begin{tcolorbox}[curristrip]
\textbf{Iter $2$:} Find every critical point of
$f(x) = x^3 - 6x^2 + 9x + 2$ on $\mathbb{R}$ and classify each as a
local maximum, local minimum, or neither.
\end{tcolorbox}
\vspace{2pt}
\begin{tcolorbox}[curristrip]
\textbf{Iter $3$:} Compute $\int_{0}^{1} e^{-x^2}\, dx$ to six decimal
places using Simpson's rule with $n = 100$ intervals, and verify with
\texttt{scipy.integrate.quad}.
\end{tcolorbox}
\end{tcolorbox}

\paragraph{Observations across the eight topic sequences.}
The selected tasks show several recurring differences. Some
later-iteration prompts combine more operations or request numerical
verification. Python verification is included in the displayed
iteration-$3$ tasks, but these selected sequences do not show when
tool use first appears during training. They also do not establish a
task-editing lineage, a general progression in difficulty, or a
causal effect of $R_{\mathrm{ref}}$.

\subsection{Full Executor Reasoning Traces}
\label{app:case:traces}

We close the appendix with four selected reasoning traces. Three use
verified problems from AIME 2024
\citep{maa2024aime1,maa2024aime2}, so the answers are independently
checkable. The fourth uses a Curriculum task from iteration~$3$. For
each case, we juxtapose the displayed anchor trace with the displayed
post-update Executor trace and include the Python block used by the
Executor for verification. These individual traces do not measure how
often the observed differences occur.

\paragraph{Trace 1: AIME 2024 I, Problem 1 (rates and times).}

\begin{tcolorbox}[frontiercase,title={Trace 1 -- AIME 2024 I Problem 1, Answer $= 204$}]
\begin{tcolorbox}[taskstrip]
\textbf{Task.} Every morning Aya goes for a $9$-kilometer walk and
stops at a coffee shop afterwards. At speed $s$ km/h the trip takes
$4$ hours including $t$ minutes at the coffee shop. At speed $s + 2$
km/h the trip takes $2$ hours $24$ minutes including the same $t$
minutes. Find the total trip time, in minutes, when Aya walks at
$s + \tfrac{1}{2}$ km/h.
\end{tcolorbox}
\vspace{2pt}
\begin{tcolorbox}[anchorstrip]
\textbf{Anchor $\pi^{(2)}$ (majority).}
Let $w_1 = 9/s$ and $w_2 = 9/(s+2)$ be walking times in hours. Then
$w_1 + t/60 = 4$ and $w_2 + t/60 = 2.4$. Subtract: $w_1 - w_2 = 1.6$.
So $9/s - 9/(s+2) = 1.6$, giving $9(s+2-s) = 1.6 s(s+2)$, i.e.
$18 = 1.6 s(s+2)$, so $s(s+2) = 11.25$, $s^2 + 2s - 11.25 = 0$,
$s = (-2 + \sqrt{4 + 45})/2 = 5/2$. Then $w_1 = 9/2.5 = 3.6$, so
$t/60 = 0.4$, giving $t = 24$ minutes. For $s + 1/2 = 3$, the walk
takes $9/3 = 3$ hours $= 180$ minutes, plus $t = 24$, so total
$\boxed{204}$ minutes.
\textcolor{currentedge!80!black}{Correct algebraically} but the
anchor's batch is split because half of the rollouts mishandle the
unit conversion of $2$ hours $24$ minutes, landing on $w_2 = 2.24$
and producing answers in the $190$--$210$ range with no consensus.
\end{tcolorbox}
\vspace{2pt}
\begin{tcolorbox}[currentstrip]
\textbf{Current $\pi^{(3)}$ (majority).}
Convert all times to hours up front: $2$ hours $24$ minutes
$= 2 + 24/60 = 2.4$. Set $u = t/60$ so $9/s + u = 4$ and
$9/(s+2) + u = 2.4$. Eliminating $u$ gives $9/s - 9/(s+2) = 1.6$, so
$s(s+2) = 9 \cdot 2 / 1.6 = 11.25$, $s = 2.5$, and $u = 0.4$ so
$t = 24$. At $s + 1/2 = 3$ km/h, walk time $= 3$ hours $= 180$
minutes; total $= 180 + 24 = \boxed{204}$.\\[2pt]
Verify with Python:\\[2pt]
\begin{tcolorbox}[pythonstrip]
from sympy import symbols, solve, Rational\\
s, t = symbols('s t', positive=True)\\
eq1 = 9/s + t/60 - 4\\
eq2 = 9/(s+2) + t/60 - Rational(12,5)\\
sol = solve([eq1, eq2], [s, t])\\
s\_val, t\_val = sol[0]\\
print(int(9/(s\_val + Rational(1,2))*60 + t\_val))  \# 204
\end{tcolorbox}
\textcolor{currentedge!80!black}{\textbf{Correct, batch unanimous}}.
\end{tcolorbox}
\end{tcolorbox}

\paragraph{Trace 2: AIME 2024 II, Problem 1 (four-set inclusion-exclusion).}

\begin{tcolorbox}[frontiercase,title={Trace 2 -- AIME 2024 II Problem 1, Answer $= 73$}]
\begin{tcolorbox}[taskstrip]
\textbf{Task.} Among $900$ residents of Aimeville, $195$ own a diamond
ring, $367$ own a set of golf clubs, $562$ own a garden spade, and
every resident owns a bag of candy hearts. There are $437$ residents
who own exactly two of these four items and $234$ who own exactly
three. Find the number of residents who own all four.
\end{tcolorbox}
\vspace{2pt}
\begin{tcolorbox}[anchorstrip]
\textbf{Anchor $\pi^{(2)}$ (majority).}
Apply three-set inclusion-exclusion to $R, G, S$ and conflate
``exactly $k$'' partition counts with $k$-fold intersection sums.
The anchor reads $437 + 234 = 671$ as the pairwise-intersection sum
and $234$ as the triple-intersection. Then
\[|R \cup G \cup S| = 195 + 367 + 562 - 671 + 234 = 687,\]
so $900 - 687 = 213$ residents own none of $R, G, S$ (i.e.\ only
candy hearts). The anchor identifies these as the all-four owners
and reports $\boxed{213}$.
\textcolor{red}{\textbf{Wrong}}: ``exactly $k$'' counts are not
$k$-fold intersection sums, so the inclusion-exclusion expression is
malformed, and the residents owning only candy hearts are not the
residents owning all four.
\end{tcolorbox}
\vspace{2pt}
\begin{tcolorbox}[currentstrip]
\textbf{Current $\pi^{(3)}$ (majority).}
Let $n_k$ be the number of residents owning exactly $k$ items, with
$n_2 = 437$, $n_3 = 234$, and $n_4 = x$ unknown. Since every resident
owns candy hearts, no one owns zero, so $n_1 + 437 + 234 + x = 900$,
giving $n_1 = 229 - x$. Now count the total membership across $R, G,
S$ only (each resident contributes $k-1$ items here, since one of
their items is always candy hearts): $0 \cdot n_1 + 1 \cdot 437 + 2
\cdot 234 + 3 \cdot x = 905 + 3x$. But this equals $|R| + |G| + |S| =
195 + 367 + 562 = 1124$. Hence $3x = 219$, $x = \boxed{73}$.\\[2pt]
Verify with Python:\\[2pt]
\begin{tcolorbox}[pythonstrip]
from sympy import symbols, solve\\
x, n1 = symbols('x n1', nonnegative=True, integer=True)\\
eq1 = n1 + 437 + 234 + x - 900\\
eq2 = 437 + 2*234 + 3*x - (195 + 367 + 562)\\
print(solve([eq1, eq2], [x, n1]))  \# \{x: 73, n1: 156\}
\end{tcolorbox}
\textcolor{currentedge!80!black}{\textbf{Correct, batch unanimous}}.
\end{tcolorbox}
\end{tcolorbox}

\paragraph{Trace 3: AIME 2024 II, Problem 3 ($2\times 3$ digit grid).}

\begin{tcolorbox}[frontiercase,title={Trace 3 -- AIME 2024 II Problem 3, Answer $= 45$}]
\begin{tcolorbox}[taskstrip]
\textbf{Task.} Find the number of ways to place a digit
($0, 1, \dots, 9$) in each of the six cells of a $2\times 3$ grid so
that (i) the sum of the two three-digit numbers formed by the two
rows is $999$, and (ii) the sum of the three two-digit numbers formed
by the three columns is $99$.
\end{tcolorbox}
\vspace{2pt}
\begin{tcolorbox}[anchorstrip]
\textbf{Anchor $\pi^{(2)}$ (majority).}
Let the cells be $a, b, c$ in the top row and $d, e, f$ in the bottom
row. Row constraint: $\overline{abc} + \overline{def} = 999$, i.e.
$100(a+d) + 10(b+e) + (c+f) = 999$. Column constraint: $\overline{ad}
+ \overline{be} + \overline{cf} = 99$, i.e. $10(a+b+c) + (d+e+f) = 99$.
The anchor concludes that every column must sum to $9$ and every
column-pair must sum to $33$, leading to $\boxed{36}$ by direct
enumeration. \textcolor{red}{\textbf{Wrong}}: the column reading
gives $10a + d$ per column, not $a + d$; the anchor incorrectly derives the
column constraint.
\end{tcolorbox}
\vspace{2pt}
\begin{tcolorbox}[currentstrip]
\textbf{Current $\pi^{(3)}$ (majority).}
With $A = a+d, B = b+e, C = c+f$, the row constraint
$100A + 10B + C = 999$ and the bounds $A, B, C \in [0, 18]$ force
$A = B = C = 9$. The column constraint
$10(a+b+c) + (d+e+f) = 99$ with $(a+b+c) + (d+e+f) = A + B + C = 27$
gives $a+b+c = 8$ and $d+e+f = 19$. Once $a+b+c = 8$ is fixed and
each digit lies in $[0,9]$, the bottom row is determined cell by cell
($d = 9-a, e = 9-b, f = 9-c$). The number of $(a,b,c)$ in $[0,9]^3$
with $a+b+c = 8$ is the stars-and-bars count $\binom{10}{2} = 45$,
since the upper bound $9$ is not active when the sum is $8$.\\[2pt]
Verify with Python:\\[2pt]
\begin{tcolorbox}[pythonstrip]
ct=0\\
for a in range(10):\\
\hspace*{1em}for b in range(10):\\
\hspace*{2em}for c in range(10):\\
\hspace*{3em}for d in range(10):\\
\hspace*{4em}for e in range(10):\\
\hspace*{5em}for f in range(10):\\
\hspace*{6em}if 100*a+10*b+c+100*d+10*e+f==999 and\\
\hspace*{8em}10*a+d+10*b+e+10*c+f==99:\\
\hspace*{7em}ct+=1\\
print(ct)  \# 45
\end{tcolorbox}
Answer $\boxed{45}$. \textcolor{currentedge!80!black}{\textbf{Correct,
batch unanimous}}.
\end{tcolorbox}
\end{tcolorbox}

\paragraph{Trace 4: A task generated by the Curriculum at iteration~$3$ (Pell equation).}

\begin{tcolorbox}[frontiercase,title={Trace 4 -- Curriculum Task at Iteration~$3$}]
\begin{tcolorbox}[taskstrip]
\textbf{Task.} Find all positive integers $n$ with $1 \le n \le 100$
such that $n^2 + (n+1)^2$ is a perfect square. (These correspond to
Pythagorean triples whose two legs are consecutive integers.)
\end{tcolorbox}
\vspace{2pt}
\begin{tcolorbox}[anchorstrip]
\textbf{Anchor $\pi^{(2)}$ (majority).}
Try small $n$ by hand. $n = 1$: $1 + 4 = 5$, not a square.
$n = 2$: $4 + 9 = 13$. $n = 3$: $9 + 16 = 25 = 5^2$ \ding{51}.
$n = 4$: $16 + 25 = 41$. $n = 5, 6, 7, 8$: $61, 85, 113, 145$.
$n = 9$: $181$. $n = 10$: $221$. None of these are squares. Conclude
$\boxed{n = 3}$ is the only solution in $[1, 100]$.
\textcolor{red}{\textbf{Wrong}}: the anchor stops after one positive
hit and never checks the rest of the range systematically, so it
misses the second solution $n = 20$.
\end{tcolorbox}
\vspace{2pt}
\begin{tcolorbox}[currentstrip]
\textbf{Current $\pi^{(3)}$ (majority).}
Set $n^2 + (n+1)^2 = k^2$, i.e. $2n^2 + 2n + 1 = k^2$. Multiply by
$2$: $(2n + 1)^2 + 1 = 2k^2$. Let $m = 2n + 1$, so $m^2 - 2k^2 = -1$,
the negative Pell equation. Its positive integer solutions are
$(m, k) = (1, 1), (7, 5), (41, 29), (239, 169), \dots$,
which give $n = (m - 1)/2 = 0, 3, 20, 119, \dots$. Excluding
$n = 0$ and capping at $100$ leaves $n \in \{3, 20\}$.
Verify with Python:\\[2pt]
\begin{tcolorbox}[pythonstrip]
import math\\
sols = []\\
for n in range(1, 101):\\
\hspace*{1em}s = n*n + (n+1)*(n+1)\\
\hspace*{1em}r = int(math.isqrt(s))\\
\hspace*{1em}if r*r == s: sols.append(n)\\
print(sols)  \# [3, 20]
\end{tcolorbox}
Verification: $3^2 + 4^2 = 25 = 5^2$, and $20^2 + 21^2 = 400 + 441
= 841 = 29^2$. Answer $\boxed{\{3,\, 20\}}$.
\textcolor{currentedge!80!black}{\textbf{Correct}}.
\end{tcolorbox}
\end{tcolorbox}

\paragraph{Observations in the four selected traces.}
Trace~1 displays a batch split in which some anchor rollouts use $2.24$
hours instead of $2.4$ for $2$h\,$24$m. Trace~2 displays confusion
between counts for exactly $k$ groups and pairwise or triple
intersection counts. Trace~3 displays an error in positional value,
where the anchor treats $\overline{ad}$ as $a + d$ rather than
$10a + d$. Trace~4 displays an incomplete search that stops at the
first positive hit. The displayed post-update Executor traces use
Python checks before reporting their final answers. These cases show
how Python is used in the selected solutions, but they do not establish
that Python caused robustness or that the pattern holds across tasks.

\FloatBarrier

\end{document}